\documentclass{article}

\usepackage{microtype}
\usepackage{graphicx}
\usepackage{subcaption}
\usepackage[dvipsnames,table]{xcolor}
\usepackage{booktabs} 

\usepackage{algorithm}
\usepackage{algorithmic}

\usepackage{hyperref}

\usepackage[preprint]{icml2026}

\usepackage{amsmath}
\usepackage{amssymb}
\usepackage{mathtools}
\usepackage{amsthm}
\usepackage{booktabs}
\usepackage{multirow}
\usepackage{adjustbox}
\usepackage{enumitem}

\usepackage[capitalize,noabbrev]{cleveref}

\theoremstyle{plain}

\theoremstyle{definition}

\theoremstyle{remark}

\usepackage{fvextra}
\DefineVerbatimEnvironment{wrapverbatim}{Verbatim}{breaklines=true,breakanywhere=true,breaksymbol={},fontsize=\small}

\usepackage[most]{tcolorbox}

\usepackage{placeins}

\usepackage[textsize=tiny]{todonotes}
\icmltitlerunning{Learning Factual Self-Verification
for Hallucination Reduction}

\newcommand{\system}{\texttt{VeriFY}}
\newcommand{\systemNM}{\texttt{VeriFY\mbox{-}NM}}
\newcommand{\systemMM}{\texttt{VeriFY\mbox{-}MM}}
\newcommand{\systemSM}{\texttt{VeriFY\mbox{-}SM}}
\newcommand{\systemKP}{\texttt{KP}}
\newcommand{\systemSC}{\texttt{SC}}

\begin{document}

\twocolumn[
  \icmltitle{Do I \textit{Really} Know?  Learning Factual Self-Verification \\ for Hallucination Reduction}



  \icmlsetsymbol{equal}{\ensuremath{*}}
   \icmlsetsymbol{senior}{\ensuremath{\dagger}}




\icmlsetsymbol{equal}{\hbox{$\diamondsuit$}}
\icmlsetsymbol{senior}{\hbox{$\spadesuit$}}

\begin{icmlauthorlist}
  \icmlauthor{Enes Altinisik}{equal,qcri}
  \icmlauthor{Masoomali Fatehkia}{equal,qcri}
  \icmlauthor{Fatih Deniz}{equal,qcri}\\
  \icmlauthor{Nadir Durrani}{senior,qcri}
  \icmlauthor{Majd Hawasly}{senior,qcri}
  \icmlauthor{Mohammad Raza}{senior,qcri}
  \icmlauthor{Husrev Taha Sencar}{senior,qcri}
\end{icmlauthorlist}

\icmlaffiliation{qcri}{
Qatar Computing Research Institute (QCRI), Hamad Bin Khalifa University (HBKU), Doha, Qatar.\\
\mbox{$^{\diamondsuit}$}\,The authors contributed equally.\quad
\mbox{$^{\spadesuit}$}\,The authors jointly supervised the work}

\icmlcorrespondingauthor{Husrev Taha Sencar}{hsencar@hbku.edu.qa}
\icmlcorrespondingauthor{Nadir Durrani}{ndurrani@hbku.edu.qa}

  \icmlkeywords{Machine Learning, ICML}

  \vskip 0.3in
]




\printAffiliationsAndNotice{}  

\begin{abstract}

Factual hallucination remains a central challenge for Large Language Models (LLMs). Existing mitigation approaches mainly rely on either external post-hoc verification or mapping uncertainty directly to abstention during fine-tuning, often resulting in overly conservative behavior. We propose \system{}, a training-time framework that teaches LLMs to reason about factual uncertainty through consistency-based self-verification. \system{} augments training with structured verification traces that guide the model to produce an initial answer, generate and answer a probing verification query, issue a consistency judgment, and then decide whether to answer or abstain. We address the challenge of reinforcing hallucinated content when training on augmented traces with a stage-level loss masking approach that excludes hallucinated answer stages from the training objective while preserving supervision over verification behavior. Across multiple model families and scales, \system{} reduces factual hallucination rates by 9.7--53.3\%, with only modest  reduction on recall (0.4--5.7\%), and generalizes across datasets when trained on a single source.\footnote{The source code, training data, and trained model checkpoints will be released upon acceptance.
}

\end{abstract}

\begin{figure}
  \centering
\includegraphics[width=0.8\linewidth]{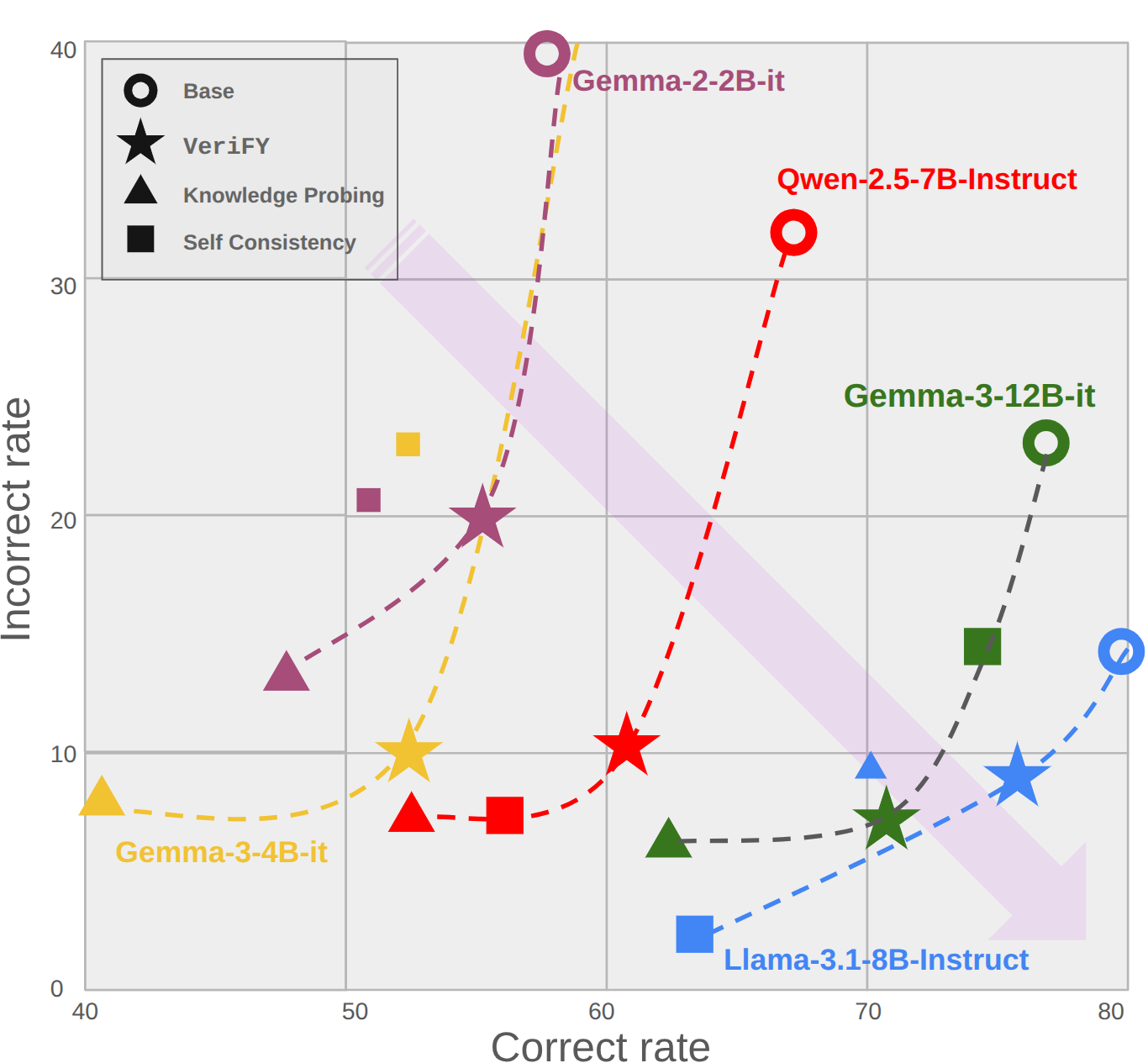}
  \caption{
Pareto frontiers of correct response rate vs. incorrect response rate for base models and their derived 
knowledge probing, self consistency, 
 and \system{} models 
for multiple model families. Knowledge probing and self consistency reduce hallucinations primarily by abstaining, which leads to a substantial drop in correct response rate. In contrast, \system{}  achieves large hallucination reductions with minor degradation to correct response rate.}
\label{fig:teaser_SM_swapped}
\vspace{-2mm}
\end{figure}

 \vspace{-4mm}
\section{Introduction}

LLMs frequently generate responses that contain incorrect or fabricated facts, a phenomenon commonly referred to as \emph{hallucination}~\cite{rawte2023troubling,10.1145/3571730}. Addressing hallucinations has become a central challenge for the reliability of LLMs. A substantial body of prior work addresses hallucinations through post-hoc intervention at inference time; including retrieval and external verification~\cite{zhang2024knowhalu,patel2025multi}, self-consistency and sampling-based checks~\cite{chen2024inside,liu2025enhancing}, reflection and revision~\cite{madaan2023self,manakul2023selfcheckgpt,dhuliawala2024chain,luo2025shaking}, and auxiliary factuality detectors~\cite{alnuhait2024factcheckmate,chang-etal-2025-monitoring,obeso2025real}. While effective in controlled settings, these approaches are extrinsic: they introduce additional complexity and latency through repeated model or external calls, and  operate only after an answer has been produced rather than at  generation, with no influence to how uncertainty is recognized or handled \cite{valentin2024cost}.


In contrast, training-time approaches aim to make hallucination mitigation intrinsic by leveraging the hypothesis that models  encode latent signals correlated with factual uncertainty~\cite{orgad2024llms,ferrando2024know,obeso2025real}.
A widely used  approach in this paradigm is refusal-style training, including methods like 
Knowledge Probing 
which explicitly train models to abstain for what they previously answered incorrectly~\cite{rtuning2024,grattafiori2024llama}. While effective at reducing hallucinations, 
such approaches lead to overly conservative models that discard a substantial fraction of  answerable queries.


In this work, we argue that bridging these two lines of work leads to more effective hallucination mitigation. Post-hoc methods show that models assessing the consistency of their own outputs can substantially reduce hallucinations, while training-time approaches expose models to internal uncertainty signals associated with incorrect generations. We propose  internalizing post-hoc verification as a learned reasoning behavior during training, enabling the model to infer sharper internal uncertainty signals that can be mapped to abstention decisions. A key challenge of such an approach is that naively training on self-verification traces risks reinforcing hallucinated content, as the model may internalize incorrect facts intended to exemplify negative inference and later refuted in a verification trace. We therefore propose a training design that disentangles learning {how to verify} from incorrect factual content.



To this end, we introduce \system{} (\textbf{Veri}fication for \textbf{F}actual Uncertaint\textbf{Y}), a framework that teaches models to effectively reason about the validity of their own responses to factoid questions through structured self-verification. 
Under \system{}'s training regime, each input query is augmented with a structured verification trace in which  an initial answer is first produced, a verification question designed to probe that answer via controlled semantic transformations (e.g., rephrasing, logical implication, or temporal variation) is generated, a verification response is produced, the original query is re-answered  in the context of the verification question-answer pair, and finally a consistency judgment is emitted to decide if the model should abstain or not. As some parts of the verification trace could contain hallucinated factual content for the sake of demonstrating the verification process (e.g. model's wrong answer to the original or verification queries), naively training on such traces risks reinforcing hallucinations. To address this challenge, \system{} incorporates stage-level loss masking, which selectively excludes hallucinated answer stages from the training objective while preserving supervision over verification, reasoning, and consistency assessment. This design enables models to internalize verification as a reasoning behavior, enabling reassessing factual associations during generation, without unintended reinforcement of false factual assertions.


Moreover, we propose a novel metric to evaluate abstention-based hallucination mitigation. Evaluation of such systems requires an explicit tradeoff between factual reliability and answer coverage: abstaining more can reduce errors, but at the cost of withholding correct answers. Standard accuracy ignores abstentions altogether, while \textit{selective accuracy}~\cite{JMLR:v11:el-yaniv10a,qazi2025scaling}, which emphasizes correctness conditional on answering, makes it difficult to reason about this tradeoff in a principled way. We propose a novel evaluation metric that measures precision and recall \textit{relative to the original model’s knowledge}. This metric enables principled comparison between hallucination mitigation strategies built on top of the same model by distinguishing beneficial from unnecessary abstentions.

Across multiple model families and scales, \system{} achieves an improved precision--recall tradeoff relative to baselines such as knowledge probing and self consistency. As  Figure~\ref{fig:teaser_SM_swapped} shows, while the other methods reduce incorrect answers primarily by abstaining leading also to a substantial drop in correct response rate, \system{} achieves balanced hallucination reductions with minimal degradation to correct responses. These gains  
demonstrate the value of verification reasoning learned at training time to model reliability. 

We make the following contributions in this work:
\textbf{i)} we introduce \system{}, a training-time hallucination mitigation framework that internalizes verification as a learned reasoning behavior through structured verification traces and stage-level loss masking, teaching models to reason about factual validity while preventing reinforcement on hallucinated content;
\textbf{ii)} we propose a novel hallucination evaluation metric that frames hallucination mitigation as a selective prediction problem, measuring precision–recall tradeoffs relative to the model’s knowledge; and
\textbf{iii)} we empirically demonstrate across multiple model families and scales that \system{} achieves an improved balance between hallucination reduction and answer coverage, without relying on external verification or inference-time intervention.

\section{Methodology}
\label{sec:methodology}


\subsection{Verification Trace Construction}
\label{sec:trace_construction}

\begin{figure}[t]
  \centering
  \includegraphics[width=\linewidth]{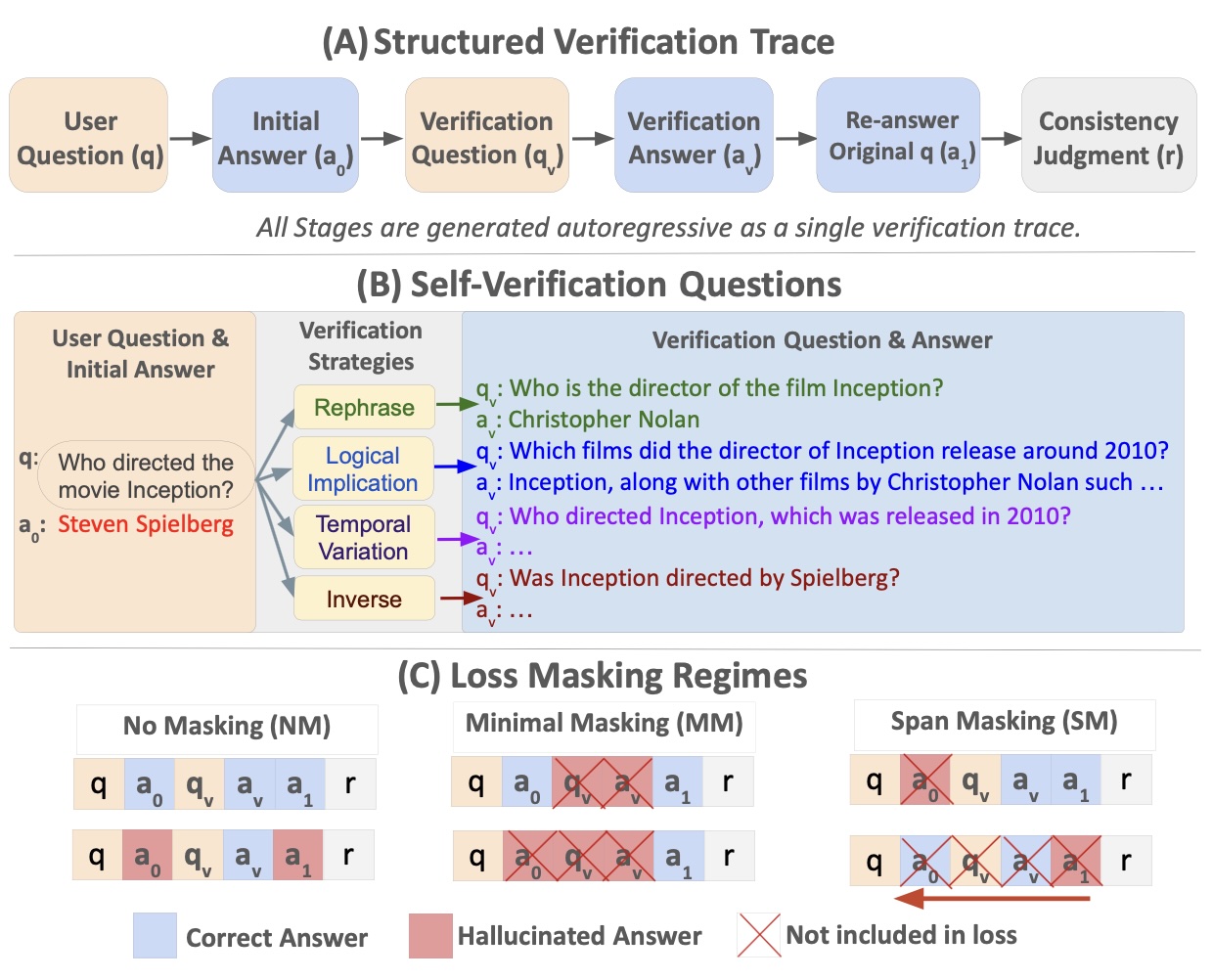} 
  \caption{Verification trace and loss masking in \system{}.
(A) A structured verification trace in which a model produces an initial answer, generates a verification question to probe factual validity, answers the verification query, re-answers the original question, and emits a consistency judgment.
(B) Verification strategies induce controlled semantic variations of the original query.
(C) Training-time loss masking regimes, illustrating how hallucinated answer stages are selectively excluded from the objective while preserving supervision over verification and consistency-judgment. \\
}
  \label{fig:verification-trace}
\vspace{-0.6mm}
\end{figure}

We define a \textit{verification trace} to be a structured representation of verification behavior that reassesses an initial response under controlled semantic perturbations. It comprises a sequence of stages with distinct functional roles (Figure \ref{fig:verification-trace}) that form the unit of supervision during training, enabling selective loss masking without suppressing verification behavior, making verification explicit and learnable. 

For a training query $q$, a verification trace is augmented as follows:
\textbf{(i)} the model produces an unconstrained initial answer $a_0$ reflecting its default generation behavior and serving as the object of verification;
\textbf{(ii)} a verification question $q_v$ is generated by a more capable model, applying a controlled semantic transformation to the original query  conditioned on $a_0$ (e.g., rephrasing or altering the semantic perspective);  \textbf{(iii)} the model produces an independent verification answer $a_v$ not conditioned on the initial response;
\textbf{(iv)} the model, conditioning on the verification pair $(q_v, a_v)$,  re-answers the original query for a revised response $a_1$; 
\textbf{(v)} a judge model emits a consistency judgment $r$ assessing whether the two responses to the original query, $a_0$ and $a_1$, are mutually supportive or inconsistent, which is used to produce a final answer or an abstention, respectively. Trace generation is performed by the same instruction-tuned model that is subsequently fine-tuned, ensuring alignment of verification behavior and the model’s intrinsic uncertainty signals.

A number of \textbf{verification strategies} are used to instantiate the verification questions, inducing controlled semantic perturbations of the original query while preserving the underlying factual content,  in order to probe the model’s confidence in its initial response. We consider a total of nine verification strategies, including rephrasing, logical implication, and temporal variation, each designed to induce a distinct semantic perturbation. Detailed descriptions and prompts for all verification strategies are provided in Appendix~\ref{sec:app:verification_strategies}.
That said, \system{}  is agnostic to the specific strategies used, allowing adaptation to the task or domain.

In this work, we instantiate \system{} on \textbf{factoid questions} with well-defined reference answers. This choice allows correctness to be evaluated unambiguously and enables explicit supervision of consistency and abstention behavior. While our experiments focus on this setting, the verification  framework itself is not inherently restricted to factoid data.

\subsection{Loss Masking Regime}
\label{sec:loss_masking}


Structured verification traces make the model’s verification behavior explicit to learn, but they may introduce hallucinated content in answer stages. Our training objective therefore needs to ensure that hallucinations are never reinforced while learning from the verification questions  and consistency judgments  even when some answers are incorrect. For this, we associated \textbf{hallucination signals}  with all answers in the training traces:  answers of the original query ($a_0, a_1$) utilized the ground-truth labels for reference, while the verification answers ($a_v$) were evaluated using the consensus of two independent judge models.\footnote{We use \textsc{Gemma-3-27B-IT} and \textsc{gpt-oss-120B} as judges.} Note that these hallucination signals are not used as direct supervision targets. 
Then, we train using \textit{stage-level loss masking}, in which 
the 
stages identified as hallucinated are excluded from the training objective. Operating at the granularity of stages avoids entangling correct and incorrect supervision within a single semantic unit so that the model never learns to reproduce false factual assertions while  learning  verification and consistency assessment in their presence.

Formally, a verification trace is represented as an ordered sequence of semantic stages
$\{s_1, \dots, s_K\}$, 
with each stage $s_j$  assigned a binary hallucination indicator $H_j \in \{0,1\}$. 
We study three loss masking regimes (Figure~\ref{fig:verification-trace}(C)):

\textbf{No Masking (NM)}
 applies supervision to the entire verification trace, without excluding any stages. All stages contribute to the training loss whether or not they contain hallucinated content. This regime serves as a baseline corresponding to standard supervised fine-tuning on verification traces, but risks reinforcing incorrect factual assertions when hallucinated answer stages are present.

\textbf{Minimal Masking (MM)}
selectively excludes only those stages identified as containing hallucinated answer content, while preserving supervision on all remaining stages. In particular, verification questions and consistency judgments remain fully supervised, even when the initial response is incorrect. 
The MM objective is defined as:
\begin{equation}
\mathcal{L}_{\mathrm{MM}}
= - \sum_{j : H_j = 0}
\sum_{t \in \mathcal{T}(s_j)}
\log p_\theta(t \mid x, s_{<j}, t_{<})
\nonumber
\end{equation}
where \(\mathcal{T}(s_j)\) denotes the set of tokens in stage \(s_j\), and \(t_{<}\) denotes the preceding tokens within that stage. Under MM, stages marked as hallucinated are excluded from the loss but remain  in the input context. This allows the model to learn how to reason about and respond to hallucinated content without being reinforced on incorrect factual assertions.

\textbf{Span Masking (SM)}
 applies a more conservative strategy by excluding an entire prefix of the verification trace that contains all hallucinations. 
Let \(k^* = \max \{ j \in \{1,\dots,K\} \mid H_j = 1 \}\) denote the index of the last stage containing hallucinated content. All stages \((s_1, \dots, s_{k^*})\) are excluded from the training loss, with the objective:
\begin{equation}
\mathcal{L}_{\mathrm{SM}}
= - \sum_{j > k^*}
\sum_{t \in \mathcal{T}(s_j)}
\log p_\theta(t \mid x, s_{<j}, t_{<})
\nonumber
\end{equation}

If hallucination occurs late in the trace (e.g., in the re-answered response), most of the trace is masked; if it occurs early (e.g., in the initial answer), relatively little supervision is removed. This regime prioritizes robustness to incorrect content at the cost of discarding additional supervision.

\section{Hallucination Reduction Metric}
\label{sec:evaluation_metrics}

Although widely used, binary correctness-based metrics are ill-suited for evaluating hallucination mitigation as they incentivize models to guess when uncertain rather than abstain~\cite{kalai2025language}. A more appropriate framing treats hallucination reduction as a selective prediction problem, where abstaining in uncertain cases is desirable. Assume the existence of an oracle that determines, for each factual question, whether the answer is \emph{known} or \emph{unknown} to a model. Under this assumption, the model’s decision to answer or abstain can be framed as prediction in binary classification: if the model attempts to answer, it is \textit{predicting} that  a question is known ($P$), leading to correct ($C$) or incorrect ($X$) response; otherwise ($N$), it abstains ($R$). This induces the following confusion matrix for a model:


\begin{itemize}[leftmargin=*, itemsep=0pt, topsep=2pt, parsep=0pt]
\item $TP$: questions predicted known by the model and are truly known, thus answered correctly ($=C$)
\item $FP$: questions predicted known by the model but are truly unknown, thus answered incorrectly ($=X$)
\item $TN$: questions predicted unknown by the model and are truly so, thus correctly abstained from ($=R^{+}$)
\item$FN$: questions predicted unknown by the model but are truly known, thus unnecessarily abstained ($=R^{-}$)
\end{itemize}

Under this framing, we could compute precision and recall:
\begin{equation}
\mathcal{P} = \frac{TP}{TP + FP}
  = \frac{\#C}{\#(C + X)}
  \label{eq:precision}
\end{equation}
\begin{equation}
\mathcal{R} = \frac{TP}{TP + FN}  = \frac{\#C}{\#(C + R^{-})}
\label{eq:recall}
\end{equation}
%
Precision answers the question: \emph{when the model chooses to answer, what fraction of answers are correct?} This corresponds to selective accuracy, a metric commonly used in hallucination mitigation~\cite{qazi2025scaling}. Recall, by contrast, measures the fraction of queries that are actually known to the model for which it chooses to answer rather than abstain.
While $C$ and $X$ can be computed directly for factual questions with gold references, the unnecessary abstentions $R^{-}$ (the count of instances that are known to the model but for which the model still conservatively abstains) are not known without an oracle (or a self-consistency--style estimation).
 For an upper bound on recall, one could assume that $R^{-} = \phi$ for base models, i.e. every abstention is done  correctly for queries the model truly does not know.

However, the complication with $R^{-}$ is resolved when comparing hallucination mitigation strategies built on top of the same model. In this setting, $R^{-}$ can be estimated by counting instances that were answered correctly by the base model but are refused after applying a mitigation intervention ($C\rightarrow R$). We then use the F1 score, the harmonic mean of precision $\mathcal{P}$ and recall $\mathcal{R}$, as our evaluation metric. This metric balances the tradeoff between correctness and coverage while accounting for the imbalance between known and unknown questions in the data. We refer to this metric as \emph{Hallucination F1}. For completeness, we also report \textit{coverage} in the appendix alongside the full results, defined as the proportion of questions that were attempted.

\begin{table*}[t]
\centering
\caption{Evaluation on the TriviaQA benchmark across small (2B--4B) and mid-sized (7B--12B) instruction-tuned models. Columns decompose model behavior into correct answers (C), hallucinated answers (X), unnecessary abstentions ($R^{-}$), and appropriate abstentions ($R^{+}$), with precision (P), recall (R), and F1 summarizing the resulting selectivity. Knowledge probing (KP) achieves high precision by aggressively increasing refusals, frequently converting correct answers into abstentions ($C \rightarrow R^{-}$). Self-consistency (SC) reduces hallucinated answers by abstaining when sampled generations disagree, improving recall over \systemKP{} but still incurring unnecessary abstentions, particularly for smaller models. \system{} instead selectively suppresses incorrect answers while retaining substantially more correct responses, yielding higher recall and F1 across all model families.}
\label{tab:evaluation_triviaqa_bigtable}
\footnotesize
\setlength{\tabcolsep}{2.5pt}
\renewcommand{\arraystretch}{0.95}
\begin{tabular}{l r|r|rr r c|m{1.1cm} | l r|r|rr r c|c}
\toprule
\multicolumn{8}{c|}{\textbf{Small-Size Models (2-4B)}} & \multicolumn{8}{|c}{\textbf{Mid-Size Models (7-12B)}} \\
\cmidrule(lr){1-8} \cmidrule(lr){9-16}
 & $C${\tiny$\uparrow$} & $X${\tiny$\downarrow$} & $R^{-}${\tiny$\downarrow$} & $R^{+}${\tiny$\uparrow$} & $\mathcal{P}${\tiny$\uparrow$} & $\mathcal{R}${\tiny$\uparrow$} & F1{\tiny$\uparrow$} &  & $C${\tiny$\uparrow$} & $X${\tiny$\downarrow$} & $R^{-}${\tiny$\downarrow$} & $R^{+}${\tiny$\uparrow$} & $\mathcal{P}${\tiny$\uparrow$} & $\mathcal{R}${\tiny$\uparrow$} & F1{\tiny$\uparrow$} \\
\midrule
\multicolumn{8}{l|}{\textbf{Gemma-2-2B}} & \multicolumn{8}{|l}{\textbf{Gemma-2-9B}} \\
\midrule
\rowcolor[gray]{0.85}\quad \textit{Base} & \textit{58.5} & \textit{40.8} & \multicolumn{2}{c}{(\textit{0.7})} & \textit{58.9} & - & - & \quad \textit{Base} & \textit{77.0} & \textit{22.0} & \multicolumn{2}{c}{(\textit{0.9})} & \textit{77.8} & - & - \\
\quad \systemKP{} & 46.6 & \textbf{13.2} & 12.5 & \textbf{27.8} & 78.0 & 78.9 & 78.5 & \quad \systemKP{} & 64.3 & 6.9 & 13.4 & \textbf{15.4} & 90.3 & 82.8 & 86.4 \\
\quad \texttt{SC} & 50.0 & 16.4 & 8.6 & 24.9 & 75.2 & 85.3 & 80.0 & \quad \texttt{SC} & 74.0 & 11.3 & \textbf{4.4} & 10.2 & 86.7 & \textbf{94.3} & 90.4 \\

\quad \systemNM{} & 49.3 & \textbf{13.2} & 10.1 & 27.4 & \textbf{78.9} & 82.9 & 80.9 & \quad \systemNM{} & 70.5 & \textbf{6.6} & 8.0 & 14.9 & \textbf{91.5} & 89.8 & 90.6 \\
\quad \systemMM{} & 54.4 & 17.5 & 6.2 & 21.9 & 75.6 & 89.7 & \textbf{82.0} & \quad \systemMM{} & 74.3 & 9.7 & 4.8 & 11.2 & 88.4 & 93.9 & 91.1 \\
\quad \systemSM{} & \textbf{55.5} & 19.6 & \textbf{5.3} & 19.6 & 73.9 & \textbf{91.2} & 81.7 & \quad \systemSM{} & \textbf{74.7} & 10.0 & 4.5 & 10.8 & 88.2 & \textbf{94.3} & \textbf{91.2} \\
\midrule
\multicolumn{8}{l|}{\textbf{Qwen2.5-3B}} & \multicolumn{8}{|l}{\textbf{Qwen2.5-7B}} \\
\midrule
\rowcolor[gray]{0.85}\quad \textit{Base} & \textit{55.5} & \textit{40.5} & \multicolumn{2}{c}{(\textit{4.0})} & \textit{57.8} & - & - & \quad \textit{Base} & \textit{67.1} & \textit{32.2} & \multicolumn{2}{c}{(\textit{0.8})} & \textit{67.6} & - & - \\
\quad \systemKP{} & 33.5 & {5.7} & 22.3 & {38.5} & {85.5} & 60.0 & 70.5 & \quad \systemKP{} & 52.9 & 7.9 & 14.1 & 25.0 & 87.0 & 78.9 & 82.8 \\
\quad \texttt{SC} & 14.8 & \textbf{1.7} & 40.6 & \textbf{42.8} & \textbf{89.6} & 26.8 & 41.2 & \quad \texttt{SC} & 54.9 & \textbf{6.9} & 12.1 & \textbf{26.0} & \textbf{88.8} & 81.9 & 85.2 \\
\quad \systemNM{} & 46.6 & 11.0 & 10.4 & 32.0 & 80.9 & 81.7 & 81.3 & \quad \systemNM{} & 59.3 & 8.9 & 8.8 & 23.0 & 86.9 & 87.1 & 87.0 \\
\quad \systemMM{} & 50.0 & 13.7 & 8.3 & 28.0 & 78.5 & 85.7 & \textbf{82.0} & \quad \systemMM{} & 60.9 & 9.8 & 7.6 & 21.7 & 86.1 & 88.9 & \textbf{87.5} \\
\quad \systemSM{} & \textbf{52.2} & 19.1 & \textbf{5.8} & 22.9 & 73.2 & \textbf{90.0} & 80.7 & \quad \systemSM{} & \textbf{61.1} & 10.5 & \textbf{7.3} & 21.0 & 85.3 & \textbf{89.3} & 87.3 \\
\midrule
\multicolumn{8}{l|}{\textbf{Llama-3.2-3B}} & \multicolumn{8}{|l}{\textbf{Llama-3.1-8B}} \\
\midrule
\rowcolor[gray]{0.85}\quad \textit{Base} & \textit{63.5} & \textit{21.4} & \multicolumn{2}{c}{(\textit{15.2})} & \textit{74.8} & - & - & \quad \textit{Base} & \textit{79.9} & \textit{14.8} & \multicolumn{2}{c}{(\textit{5.3})} & \textit{84.4} & - & - \\
\quad \systemKP{} & 54.3 & 11.4 & 10.1 & 24.2 & 82.7 & 84.3 & 83.5 & \quad \systemKP{} & 70.2 & 9.5 & 8.8 & 11.5 & 88.1 & 88.9 & 88.5 \\
\quad \texttt{SC} & 32.1 & \textbf{1.7} & 31.4 & \textbf{34.9} & \textbf{95.0} & 50.5 & 66.0 & \quad \texttt{SC} & 58.4 & \textbf{2.1} & 21.5 & \textbf{18.0} & \textbf{96.6} & 73.1 & 83.2 \\
\quad \systemNM{} & 60.5 & 14.9 & \textbf{6.1} & 18.5 & 80.2 & 90.8 & 85.2 & \quad \systemNM{} & 73.8 & 10.4 & 6.7 & 9.1 & 87.6 & 91.7 & 89.6 \\
\quad \systemMM{} & 62.3 & 12.4 & 6.3 & 18.9 & 83.4 & 90.8 & \textbf{86.9} & \quad \systemMM{} & 75.0 & 10.1 & 6.1 & 8.9 & 88.2 & 92.5 & 90.3 \\
\quad \systemSM{} & \textbf{62.4} & 12.7 & 6.3 & 18.6 & 83.1 & \textbf{90.9} & 86.8 & \quad \systemSM{} & \textbf{75.8} & 9.7 & \textbf{5.3} & 9.2 & 88.7 & \textbf{93.5} & \textbf{91.0} \\
\midrule
\multicolumn{8}{l|}{\textbf{Gemma-3-4B}} & \multicolumn{8}{|l}{\textbf{Gemma-3-12B}} \\
\midrule
\rowcolor[gray]{0.85}\quad \textit{Base} & \textit{57.1} & \textit{42.8} & \multicolumn{2}{c}{(\textit{0.1})} & \textit{57.1} & - & - & \quad \textit{Base} & \textit{76.8} & \textit{23.1} & \multicolumn{2}{c}{(\textit{0.1})} & \textit{76.9} & - & - \\
\quad \systemKP{} & 41.1 & 8.1 & 16.4 & 34.4 & 83.5 & 71.6 & 77.1 & \quad \systemKP{} & 62.4 & \textbf{6.4} & 14.5 & \textbf{16.8} & 90.7 & 81.2 & 85.7 \\
\quad \texttt{SC} & \textbf{52.7} & 22.5 & \textbf{4.4} & 20.5 & 70.1 & \textbf{92.3} & 79.7 & \quad \texttt{SC} & \textbf{72.8} & 13.7 & \textbf{4.0} & 9.5 & 84.2 & \textbf{94.8} & 89.2 \\
\quad \systemNM{} & 44.1 & \textbf{6.7} & 14.5 & \textbf{34.7} & \textbf{86.8} & 75.2 & 80.6 & \quad \systemNM{} & 70.8 & 10.1 & 6.5 & 12.6 & 87.5 & 91.5 & 89.5 \\
\quad \systemMM{} & 52.5 & 14.1 & 8.3 & 25.0 & 78.8 & 86.3 & 82.4 & \quad \systemMM{} & 72.0 & 8.8 & 6.4 & 12.8 & 89.1 & 91.8 & \textbf{90.5} \\
\quad \systemSM{} & 51.7 & 10.4 & 9.2 & 28.7 & 83.2 & 85.0 & \textbf{84.1} & \quad \systemSM{} & 70.9 & 7.2 & 7.7 & 14.2 & \textbf{90.8} & 90.2 & \textbf{90.5} \\
\bottomrule
\end{tabular}
\end{table*}

\section{Evaluation}

\subsection{Experimental Setup}
\label{sec:setup}


\textbf{Datasets:} We construct training data by bootstrapping structured verification traces from factual questions drawn from  TriviaQA dataset~\citep{joshi-etal-2017-triviaqa}.\footnote{\url{https://nlp.cs.washington.edu/triviaqa}}
 TriviaQA consists of factoid-style trivia questions paired with reference answers, making it well suited for eliciting factual errors and hallucinations in language models. We use the  training split (87,622 questions) for supervised fine-tuning, and reserve the validation and test splits for in-domain evaluation. Cross-dataset generalization is assessed using HotpotQA~\cite{dataset:yang2018hotpotqa}, NQ-Open~\cite{dataset:nqopen:kwiatkowski2019natural}, and the Nov~2025 release of FreshQA~\cite{dataset:freshqa:vu2024freshllms}.

\textbf{Trace assembly and filtering:}
For trace construction, all generated stages are assembled into a single structured verification trace per training question  enclosed within explicit delimiters (\texttt{<verify>} \texttt{</verify>}) to capture internal deliberation. A short final response, indicating either an answer or an abstention is appended outside the delimiters and treated as the surface-level output.  To ensure a consistent and unambiguous supervision signal, we retain only traces that exhibit aligned correctness and consistency outcomes: either the initial response is correct and the revised response is judged consistent, or the initial response is incorrect and the revised response is judged inconsistent. Traces that violate these conditions are discarded to avoid conflicting supervision. 
Example traces 
are provided in Appendix~\ref{appendix:subsec:traces_examples}.

\textbf{Models:}
\label{sec:models}
We evaluate \system{} across multiple instruction-tuned language model families and parameter scales to assess robustness across architectures and capacities. Our experiments include models from Gemma (Gemma-2 and Gemma-3) \cite{team2025gemma}, LLaMA (3.1 and 3.2), \cite{grattafiori2024llama} and Qwen2.5 \cite{yang2024qwen25} families, with sizes between 2B and  9B parameters. All models are initialized from publicly released instruction-tuned checkpoints, reflecting realistic deployment settings in which models already exhibit basic alignment and refusal behavior. To ensure alignment between learned verification behavior and intrinsic uncertainty, verification traces are generated separately for each  model 
with all answer stages produced by the  model itself, while verification questions and consistency judgments are generated by a higher-capacity model (Gemma-3-27B-IT), enabling reliable probing while remaining agnostic to the specific capabilities of the student model.
Complete details of the training setup and optimization parameters are provided in Appendix~\ref{sec:appendix_training_parameters}.


\textbf{Baselines:} We report results for \textbf{i)} knowledge probing (\systemKP{}), a training-time method \cite{grattafiori2024llama}; and \textbf{ii)} self-consistency (\systemSC{}-$k$), a post-hoc approach that produces $k$ generations sampled at temperature $1.0$. 
See Appendix \ref{sec:appendix_baselines} for details.


\subsection{Results}
\label{sec:experiments}

We report in Table~\ref{tab:evaluation_triviaqa_bigtable} detailed statistics that decompose model behavior into correct answers $C$, incorrect answers $X$, unnecessary abstentions $R^{-}$, and appropriate abstentions $R^{+}$, together with derived precision $\mathcal{P}$, recall $\mathcal{R}$, and F1 scores. This decomposition allows us to analyze both \textit{when}  and \emph{how} hallucinations are reduced.  

\textbf{Comparing against the baselines:} Table~\ref{tab:evaluation_triviaqa_bigtable} reports results on the TriviaQA benchmark for small (2--4B) and mid-sized (7--12B) models. Across all model families, \system{} alters the error profile of the base models in a consistent and principled way.
\system{} selectively suppresses incorrect answers while preserving a substantially larger fraction of correct ones.
As a result, \system{} achieves lower unnecessary abstention rates ($R^{-}$) and higher recall than the base models at comparable or higher precision, yielding improved F1 scores.
This indicates that verification-based training reduces hallucinations primarily by eliminating incorrect answers rather than by discarding answerable queries. 

In contrast, knowledge probing (KP)  reduces hallucinations by aggressively increasing abstentions. While this leads to large reductions in incorrect answers (X), it also produces a marked rise in unnecessary abstentions ($R^{-}$), frequently from previously correct base-model answers.
Consequently, recall drops sharply under \systemKP{}, particularly for smaller models, revealing an overly conservative failure mode that discards non-trivial model knowledge.
Self-consistency (\systemSC{}) achieves higher recall and F1 by refusing fewer correct answers than \texttt{KP}. However, it relies on disagreement across sampled generations at inference time to decide when to abstain, which leads to unnecessary abstentions and lower recall for smaller models. 
In practice, we observe that smaller Qwen and LLaMA models frequently fail to reliably check  consistency, suggesting that effective deployment of \systemSC{} may require a more capable external judging model.
In contrast, \system{} reduces incorrect answers more selectively by learning consistency checks during training, maintaining higher recall and F1 without inference-time sampling.

\textbf{Comparing loss-masking regimes:} Training with full structured verification traces (\systemNM{}) already recovers a significant portion of the recall lost under \systemKP{}, indicating that explicit self-verification enables models to better distinguish genuinely uncertain cases from answerable ones. Nevertheless, \systemNM{} exhibits elevated unnecessary abstention rates, particularly for smaller models, suggesting that unmasked supervision does reinforce hallucinated content during training. Introducing stage-level loss masking leads to the strongest results. Both \systemMM{} and \systemSM{} consistently outperform \systemNM{} across all model families.
By preventing reinforcement on hallucinated stages while preserving supervision over verification and consistency reasoning, we reduce incorrect answers while minimizing unnecessary abstentions. Between the two masking strategies, a stable tradeoff emerges: Span Masking (SM) typically achieves slightly higher recall, while Minimal Masking (MM) attains marginally higher precision. Importantly, both approaches maintain substantially higher recall than \texttt{KP} with comparable or lower hallucination rates.

\textbf{Cross-Dataset Generalization:}   Table~\ref{tab:generalization_results} reports generalization performance using Gemma-2-2B and Gemma-2-9B backbones across three out-of-domain benchmarks: HotpotQA, NQ-Open, and FreshQA. Despite being trained only on TriviaQA, \system{} consistently reduces hallucination rates relative to the base models, \systemKP{}, and \systemSC{} across datasets, while maintaining substantially higher answer recall. This trend holds across model scales. Notably, \systemKP{} often attains high precision but suffers a sharp degradation in recall under distribution shift, indicating brittle abstention behavior that does not generalize beyond the training distribution. \systemSC{} improves recall relative to \systemKP{} but does so at the cost of lower precision. 
In contrast, \systemSM{} and \systemMM{} training preserves significantly higher recall while still lowering hallucination rates. This suggests that the learned verification behavior is not dataset-specific, but  reflects more generalizable reasoning about factual uncertainty. 
Results on the world-knowledge benchmark MMLU further show that our training does not materially degrade models’ parametric knowledge, as summarized in Table~\ref{tab:benchmarks} (Appendix~\ref{sec:app:benchark}).

Across both in-domain and cross-dataset evaluations, 
\system{} consistently reduces hallucinated outputs while preserving answer coverage. Verification traces alone encourage explicit reasoning about factual uncertainty, but the strongest and most robust gains arise when they are paired with stage-level loss masking,
enabling \system{} to maintain high recall under distribution shift without sacrificing precision. Together, \systemSM{} and \systemMM{} achieve the best trade-off between hallucination reduction and coverage, outperforming prior approaches across model scales and evaluation settings.
Complete results for all models and benchmarks are provided in Appendix~\ref{sec:appendix_generalization_results}.

\begin{table}
\centering
\caption{Cross-dataset evaluation on HotpotQA, NQ-Open, and FreshQA. Models trained on TriviaQA are evaluated under distribution shift using Gemma-2-2B and Gemma-2-9B backbones. 
}
\label{tab:generalization_results}
\adjustbox{max width=\columnwidth}{
\begin{tabular}{@{}lrrr rrr rrr@{}}
\toprule
 & \multicolumn{3}{c}{\textbf{HotpotQA}} & \multicolumn{3}{c}{\textbf{NQOpen}} & \multicolumn{3}{c}{\textbf{FreshQA}} \\
\cmidrule(lr){2-4} \cmidrule(lr){5-7} \cmidrule(lr){8-10}
 & $\mathcal{P}${\tiny$\uparrow$} & $\mathcal{R}${\tiny$\uparrow$} & F1{\tiny$\uparrow$} & $\mathcal{P}${\tiny$\uparrow$} & $\mathcal{R}${\tiny$\uparrow$} & F1{\tiny$\uparrow$} & $\mathcal{P}${\tiny$\uparrow$} & $\mathcal{R}${\tiny$\uparrow$} & F1{\tiny$\uparrow$} \\
\midrule
\multicolumn{9}{l}{\textbf{Gemma-2-2B}} \\

\midrule
\rowcolor[gray]{0.85} \quad \textit{Base} & \textit{31.4} & - & - & \textit{40.4} & - & - & \textit{37.3} & - & - \\
\quad \systemKP{} & \textbf{51.7} & 43.0 & 47.0 & \textbf{57.1} & 57.6 & 57.3 & \textbf{54.1} & 69.4 & 60.8 \\
\quad \systemSC{} & 47.5	& 58.8	& 52.6 & 53.8 &	71.7 &	61.5 & 46.6 & \textbf{82.7} & 59.6 \\
\quad \systemNM{} & 48.8 & 59.9 & \textbf{53.8} & 52.3 & 74.9 & \textbf{61.6} & 46.1 & 68.6 & 55.1 \\
\quad \systemSM{} & 40.4 & \textbf{77.9} & 53.2 & 48.7 & \textbf{86.0} & 62.2 & 45.2 & 78.6 & 57.4 \\
\quad \systemMM{} & 42.4 & 73.6 & \textbf{53.8} & 49.6 & 80.8 & {61.5} & 49.3 & 81.8 & \textbf{61.5} \\
\midrule
\multicolumn{9}{l}{\textbf{Gemma-2-9B}} \\
\midrule
\rowcolor[gray]{0.85} \quad \textit{Base} & \textit{44.2} & - & - & \textit{53.5} & - & - & \textit{50.0} & - & - \\
\quad \systemKP{} & \textbf{64.9} & 42.4 & 51.3 & \textbf{70.5} & 60.2 & 64.9 & \textbf{67.2} & 61.2 & 64.1 \\
\quad \systemSC{} & 55.9 & 79.3 &	65.6 & 62.3 & 84.8 & 71.8 & 54.8 & \textbf{90.9} & 68.4 \\
\quad \systemNM{} & 61.8 & 69.6 & 65.5 & 66.6 & 82.9 & 73.8 & 59.5 & 82.2 & 69.0 \\
\quad \systemSM{} & 54.3 & \textbf{81.9} & 65.3 & 62.2 & \textbf{88.8} & 73.2 & 56.4 & 84.1 & 67.5 \\
\quad \systemMM{} & 56.0 & 80.7 & \textbf{66.1} & 64.3 & 86.9 & \textbf{73.9} & 58.2 & 87.7 & \textbf{69.9} \\
\bottomrule
\end{tabular}
}
\vspace{-3mm}
\end{table}

\subsection{Ablation and Component Analysis}
We analyze the contribution of individual components of \system{} to identify which aspects of the verification process are essential and clarify the settings in which the approach is effective. 
Specifically, we examine three factors: (i) the choice and diversity of verification strategies, (ii) whether inference-time perturbations can substitute for learned verification behavior, and (iii) the extent to which generated verification traces transfer across model families. 



\textbf{Which verification strategies matter?}  
We begin by comparing models trained with traces that use a \emph{single} verification strategy (Appendix~\ref{sec:app:verification_strategies}). As shown in Table~\ref{tab:ablation_individual_strategies}, no individual strategy consistently dominates across all metrics. Rephrasing achieves the highest F1 scores overall, primarily due to stronger recall, while implication-based and inverse-query strategies tend to favor precision at the expense of recall. This variation indicates that different strategies probe complementary aspects of factual uncertainty. 
Importantly, the relative ranking of strategies is consistent across model scales, suggesting that their effectiveness is not strongly dependent on model size. This stability motivates combining multiple strategies rather than committing to one.

\begin{table}
\centering
\caption{Single-strategy \system{} on TriviaQA using Gemma-2-2B and Gemma-2-9B. No individual strategy dominates across metrics, with rephrasing favoring recall while implication-based and inverse-query strategies favoring precision.}
\label{tab:ablation_individual_strategies}
\small
\setlength{\tabcolsep}{4pt}
\renewcommand{\arraystretch}{1.0}
\adjustbox{max width=\columnwidth}{
\begin{tabular}{@{}lrrr@{\hspace{0.8em}}|@{\hspace{0.8em}}rrr@{}}
\toprule
\multirow{2}{*}{\textbf{\shortstack{Verification\\Strategy}}} & \multicolumn{3}{c}{\textbf{Gemma-2-2B}} & \multicolumn{3}{c}{\textbf{Gemma-2-9B}} \\
\cmidrule(lr){2-4} \cmidrule(lr){5-7}
 & $\mathcal{P}${\tiny$\uparrow$} & $\mathcal{R}${\tiny$\uparrow$} & F1{\tiny$\uparrow$} & $\mathcal{P}${\tiny$\uparrow$} & $\mathcal{R}${\tiny$\uparrow$} & F1{\tiny$\uparrow$} \\
\midrule
Rephrase & 89.91 & \textbf{73.95} & \textbf{81.15} & 96.07 & \textbf{86.81} & \textbf{91.21} \\
Comparative & 91.03 & 71.78 & 80.27 & 95.49 & 86.14 & 90.57 \\
Temporal & 90.98 & 71.17 & 79.86 & 96.43 & 84.64 & 90.15 \\
Disjunction & 89.94 & 72.00 & 79.98 & 95.81 & 84.77 & 89.95 \\
Inverse & 91.49 & 70.34 & 79.53 & \textbf{96.71} & 84.06 & 89.94 \\
Abstraction & 91.66 & 70.20 & 79.51 & 96.48 & 84.44 & 90.06 \\
Implication & \textbf{92.53} & 69.09 & 79.11 & 96.19 & 84.89 & 90.19 \\
Authority & 90.72 & 69.48 & 78.69 & 93.79 & 86.43 & 89.96 \\
Justification & 91.13 & 68.70 & 78.34 & 96.63 & 84.77 & 90.31 \\
\bottomrule
\end{tabular}
}
\end{table}

\begin{figure}
  \centering
  \includegraphics[width=0.77\linewidth]{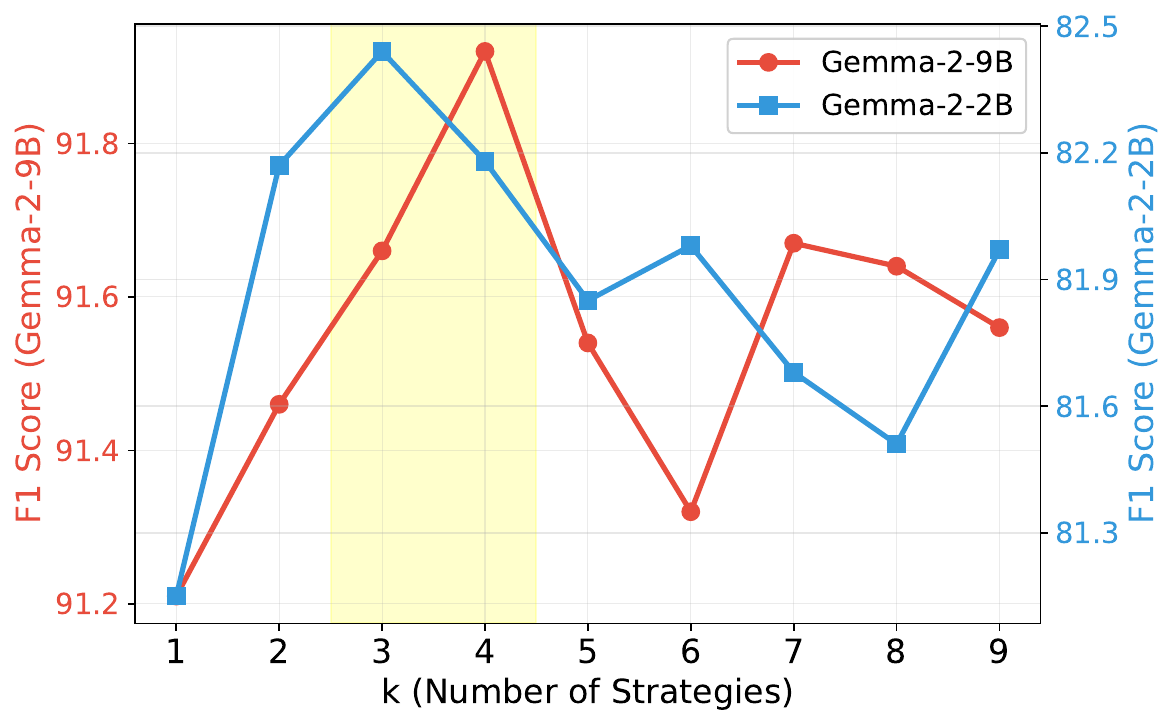} 
  \caption{Effect of the number of distinct verification strategies used for the traces on F1 performance of \system{}.
  For both Gemma-2-2B and Gemma-2-9B, performance improves as complementary strategies are added, with diminishing returns and mild degradation beyond 4, suggesting increased redundancy and noise.
 }
\label{fig:verification_strategy_count}
\vspace{-3mm}
\end{figure}

Figure~\ref{fig:verification_strategy_count} illustrates the effect of incrementally combining multiple strategies, ordered by F1, when creating training traces. Performance initially improves as additional strategies are introduced, then a diminishing return could be observed beyond four strategies. Extra strategies appear to introduce redundancy and noise into the verification trace dataset. Based on this observation, we use the top four strategies in all subsequent experiments. Crucially, \system{} is strategy-agnostic; domain- or task-specific policies can be incorporated without modifying the training objective.

\textbf{Can inference-time perturbations replace learned self-verification?}
We study inference-time contextual \textit{shaking}: a prompt-level intervention that induces stochastic drift during generation, without modifying model parameters or decoding. This analysis tests whether inference-time perturbations can serve as a reliable uncertainty-mitigation mechanism. To induce shaking, we modify the chat template by removing the \texttt{end\_of\_turn} marker from the assistant response, forcing the model to continue generation from previously produced text. 
We then prepend an auxiliary question--answer pair $({q'}, {a'})$, randomly drawn from the same dataset, before instructing the model to answer the original query.
We also evaluate a variant in which the prompt includes an explicit draft response $a_0$ to the original query prior to the perturbation (denoted $a_0$+Pert). Exact prompt templates are provided in Appendix~\ref{app:shaking_templates}. No decoding parameters or model weights are modified in any setting.

Table~\ref{tab:inference_shaking_ablation} shows that inference-time perturbations consistently degrades recall (i.e., coverage of correct responses) and F1 across both base and \systemKP{} models. Moreover, combining perturbations with \systemKP{} further amplifies recall loss, indicating that inference-time contextual perturbations interact poorly with abstention-based mechanisms. In contrast, \system{} achieves higher recall and F1, supporting the conclusion that uncertainty reasoning must be learned during training rather than imposed at inference.

\begin{table}
\centering
\caption{Performance of prompt-level contextual perturbations (Pert) and draft-response perturbations ($a_0$+Pert), compared against base models, knowledge probing (KP), and \system{}. Inference-time perturbations lead to lower recall and F1 across models, while \system{} maintains higher recall and performance.
}
\label{tab:inference_shaking_ablation}
\small
\begin{tabular}{@{}lrrr@{\hspace{0.8em}}|@{\hspace{0.8em}}rrr@{}}
\toprule
 & \multicolumn{3}{c}{\textbf{Gemma-2-2B}} & \multicolumn{3}{c}{\textbf{Gemma-2-9B}} \\
\cmidrule(lr){2-4} \cmidrule(l){5-7}
\textbf{Shaking Strategy} & $\mathcal{P}${\tiny$\uparrow$} & $\mathcal{R}${\tiny$\uparrow$} & F1{\tiny$\uparrow$} & $\mathcal{P}${\tiny$\uparrow$} & $\mathcal{R}${\tiny$\uparrow$} & F1{\tiny$\uparrow$} \\
\midrule
\rowcolor[gray]{0.85} \textit{Base} & \textit{58.9} & - & - & \textit{77.8} & - & - \\
Base + Pert & \textbf{94.3} & 56.4 & 70.6 & 92.1 & 76.9 & 83.8 \\
Base + $a_0$ + Pert & 79.9 & 69.8 & 74.5 & \textbf{99.7} & 77.9 & 87.4 \\
\systemKP{} + Pert & 92.5 & 57.4 & 70.8 & 76.3 & 85.3 & 80.5 \\
\systemKP{} + $a_0$ + Pert & 79.9 & 69.8 & 74.5 & 59.3 & 88.2 & 70.9 \\
\midrule
\system{} & 73.9 & \textbf{91.2} & \textbf{81.7} & 88.2 & \textbf{94.3} & \textbf{91.2} \\
\bottomrule
\end{tabular}
\vspace{-4mm}
\end{table}

\textbf{Does verification behavior transfer across models?}
In all experiments presented so far, verification traces are generated by the model itself. Thus, a practical concern for a training-time approach is whether data generated for one model can be reused to train another. Here, we focus specifically on whether {uncertainty and abstention signals} transfer across models. 
This experiment trains a model using traces generated by a different model, which implicitly introduces external knowledge the trained model might not have. Since our primary interest is the transfer of abstention behavior across models, we explicitly report  $R$ (recall) in Table~\ref{tab:intramodel_performance}.

As the results show, a key observation is that recall remains extremely high under inter-family transfer. While this suggests that transferred verification data encourages models to continue answering questions rather than abstaining, it also reveals an important limitation: \emph{inter-family traces fail to reliably teach calibrated abstention behavior}. This manifests as increased false positives and reduced precision, indicating that the model answers aggressively even in cases where abstention would be appropriate. We hypothesize that this failure arises from family-specific uncertainty representations embedded within the model’s internal dimensions. When transferred across families, these signals no longer align cleanly with the target model’s representations, resulting in weakened or distorted abstention cues.

Overall, these results indicate that 
effective transfer of uncertainty reasoning requires alignment at the architectural and representational level, which is naturally satisfied within a model family but largely absent across families. This finding reinforces the importance of same-family or higher-capacity trace generation when training \system{} for reliable hallucination mitigation.

\section{Related Work}


\textbf{Training-time intrinsic hallucination mitigation:}
Recent work suggests that hallucinations are anticipated by intrinsic signals encoded in a model’s internal representations rather than arising purely at the output level. \citet{orgad2024llms} show that LLMs often internally encode factual correctness, even localizing truth signals to specific tokens, and may represent correct answers despite consistently hallucinated outputs. Similarly, \citet{ferrando2024know} demonstrate that models possess internal mechanisms for recognizing whether an entity is known or unknown, which causally influence hallucination and refusal behavior, while \citet{obeso2025real} show that hallucinated entities in long-form generation can be detected in real time via lightweight probes over intermediate activations.
Separately, a class of training-time methods, including knowledge probing \citep{grattafiori2024llama} and R-Tuning \citep{rtuning2024}, aims to reduce hallucinations by converting incorrect responses into abstention-style supervision during fine-tuning, thereby encouraging models to answer only when sufficiently confident. While not originally derived from analyses of intrinsic uncertainty representations, these approaches can be understood as implicitly leveraging such signals by collapsing uncertainty into refusal under clean supervision. In contrast, our work trains models to explicitly reason about factual validity at training time, introducing the challenge of learning from self-generated verification traces that may themselves contain hallucinated content--a problem we address through 
verification with stage-level loss masking.

\begin{table}[]
\centering
\caption{Inter- and intra-family verification data transferability. Each row reports performance when a model is trained using verification traces generated by another model. While answer generation transfers reasonably well, inter-family transfer exhibits inflated recall but reduced precision, indicating a failure to transfer calibrated abstention behavior.}
\label{tab:intramodel_performance}
\small
\setlength{\tabcolsep}{4pt}
\adjustbox{max width=\columnwidth}{
\begin{tabular}{@{}llrrrrrr@{}}
\toprule
\textbf{Training Data} & \textbf{Model} & $R^{-}${\tiny$\downarrow$} & $R^{+}${\tiny$\uparrow$} & $\mathcal{P}${\tiny$\uparrow$} & $\mathcal{R}${\tiny$\uparrow$} & F1{\tiny$\uparrow$} \\
\midrule
Qwen2.5-3B & Qwen2.5-7B & 6.9 & 17.8 & 82.2 & 90.0 & 85.9 \\
 & Gemma-2-2B & 0.7 & 3.4 & 62.6 & 98.8 & 76.7 \\
\midrule
Qwen2.5-7B & Gemma-2-9B & 0.2 & 1.3 & 79.9 & 99.7 & 88.7 \\
 & Qwen2.5-3B & 3.1 & 12.3 & 69.5 & 95.0 & 80.2 \\
\midrule
Gemma-2-2B & Gemma-2-9B & 6.6 & 10.2 & 85.6 & 91.5 & 88.5 \\
 & Qwen2.5-3B & 0.0 & 4.0 & 57.8 & 100.0 & 73.2 \\
\midrule
Gemma-2-9B & Qwen2.5-7B & 0.0 & 0.3 & 69.2 & 100.0 & 81.8 \\
 & Gemma-2-2B & 1.2 & 3.1 & 66.6 & 98.1 & 79.4 \\
\bottomrule
\end{tabular}
}
\vspace{-4mm}
\end{table}

\textbf{Verification and reasoning for hallucination mitigation.}
A broad class of approaches mitigates hallucinations through verification and reasoning at inference time, either via external evidence or model-centric self-consistency. External verification methods such as SAFE~\citep{wei2024long}, FactScore~\citep{min2023factscore}, FacTool~\citep{chern2023factool}, and FactCheck~\citep{shami2025fact} decompose generated text into atomic claims and evaluate them against retrieved evidence using auxiliary models or LLM-based judges, while related techniques like SELF-REFINE~\citep{madaan2023self} rely on iterative self-feedback without external sources. Model-centric inference-time methods encourage self-verification through internal reasoning or agreement, including RARR~\citep{gao2023rarr}, which revises answers by reasoning over retrieved evidence, SelfCheckGPT~\citep{manakul2023selfcheckgpt}, which detects hallucinations via inconsistency across sampled generations, and self-consistency-based aggregation methods that treat stability across reasoning paths as a proxy for correctness~\citep{wang2022self,chen2024inside}. Among these, Chain-of-Verification (CoVe)~\citep{dhuliawala2024chain} introduces a structured draft–verify–revise procedure that explicitly decomposes and checks claims before final generation. While effective at reducing hallucinations, these methods operate entirely at inference time and incur additional computational cost through multiple model calls, retrieval steps, or sampling passes. In contrast, our approach internalizes verification as a learned behavior during training, using structured verification traces and stage-level loss masking to prevent hallucinated content from being reinforced.

\section{Conclusion}
We presented \system{}, a training-time framework for reducing factual hallucinations by teaching language models to explicitly reason about the validity of their own answers. By supervising structured self-verification traces and masking hallucinated stages during training, \system{} reduces incorrect generations without relying on abstention-only behavior or inference-time intervention. 
Across multiple model families, this approach achieves substantial hallucination reduction while largely preserving answer coverage, and generalizes across datasets when trained on a single source. 
While we focuses on factoid questions, the proposed framework is not inherently tied to this setting. Overall, learning verification behavior during training provides a principled alternative to refusal-centric hallucination mitigation.


\section*{Impact Statement}
This work aims to improve the factual reliability of language models by reducing hallucinated outputs, and consequently increase the trustworthiness of AI-assisted systems in contexts where factual accuracy is important. By focusing on training-time hallucination mitigation learning rather than relying on inference-time intervention or broad abstention, the approach is intended to reduce incorrect generations while preserving useful model behavior. As with prior work on hallucination mitigation, the societal impact of these techniques depends on how they are integrated into real-world systems, with oversight, evaluation, and deployment constraints having a paramount importance. We do not anticipate significant negative societal impacts of this work beyond those already associated with the deployment of LLMs.


    

\bibliography{custom}
\bibliographystyle{icml2026}

\appendix
\onecolumn



\section{Data Generation}

\subsection{Verification Strategies}
\label{sec:app:verification_strategies}

Below, we describe the verification strategies used in this work. For illustration, we use the example question, $q$:~‘What is the capital of Canada?’ and its wrong original answer 'Toronto' to show the corresponding verification question $q^{(v)}$ generated by each strategy.

\begin{itemize}
    \item \textbf{Rephrasing:}  Reformulate the original question to query the same fact from a different linguistic or semantic perspective, without introducing new information. \emph{Example:} “What Canadian city serves as the official seat of the national government?” 
    \item \textbf{Implication:} Generates a question about a fact that would logically follow if the model’s initial answer were correct, probing whether those implications hold. \emph{Example:} “If Toronto were the capital of Canada, where would you expect to find the Canadian Parliament and federal government offices?”
    \item \textbf{Inverse:} Generates a reverse-direction query that asks the model to recover the original fact from its initial answer. \emph{Example:} “Toronto is the capital of which country?”
    \item \textbf{Justification:} Prompts the model to explain or justify its initial answer, revealing whether the answer is supported by a coherent factual rationale. \emph{Example:} “Why do you say that Toronto is the capital of Canada? What makes it the capital?”
    \item \textbf{Temporal / Causal Probing:} Queries the historical timeline or causal process underlying the asserted fact, testing whether the answer is supported by a coherent sequence of events. \emph{Example:} “When and why was Toronto chosen as the capital of Canada?”
    \item \textbf{Abstraction / Categorization Probing:} Queries the role, category, or functional status of the proposed answer within a broader conceptual framework. \emph{Example:} “What role does Toronto play in Canada’s political system—is it a provincial capital, the federal capital, or something else?”
    \item \textbf{Mutual Exclusivity / Disjunction Probing:} Introduces a mutually exclusive alternative that forces the model to reconcile conflicting factual commitments. \emph{Example:} “If Toronto were the capital of Canada, which city would host the Canadian Parliament?”
    \item \textbf{Analogical or Comparative Probing:} Poses a parallel question in a related context to assess whether the model applies the same reasoning consistently across analogous cases. \emph{Example:} “If Toronto were the capital of Canada because it is the largest and most well-known city, would the same reasoning imply that New York City is the capital of the United States?”
    \item \textbf{Source or Authority Probing:} Asks where the stated fact would be documented or how it could be verified using authoritative references. \emph{Example:} “Where could one find an official source confirming that Toronto is the capital of Canada, such as a government publication or encyclopedia?” 
\end{itemize}

\subsection{Verification Trace Examples}
\label{appendix:subsec:traces_examples}


Examples of verification traces generated for supervised training.

\begin{tcolorbox}[title={\textbf{Example 1:} $a_0$ is Correct and $a_1$ is Consistent; In this case the model produces a final response since its revised response is consistent with its original response. Model's correct answer is in \textcolor{ForestGreen}{green} while the final model response is in \textcolor{cyan}{blue}.}, colback=red!5!white]

\textbf{Question ($q$):} Who won Super Bowl XX? 

\textbf{$<$verify$>$} \newline
Let me first try to answer the question. My answer: \textcolor{ForestGreen}{The **Chicago Bears** won Super Bowl XX.} \newline

Now, let me try to verify my answer by asking a followup verification question. Given the question, my verification strategy is: **Rephrase Questions** \newline

*Verification Question*: Which team emerged victorious in Super Bowl XX?

*My response to the Verification Question*: \textcolor{ForestGreen}{The **Chicago Bears** won Super Bowl XX.} \newline

Now based on the above verification, let me try to answer the question again. My answer based on the verification:

\textcolor{ForestGreen}{The Chicago Bears.} \newline

Now let me check if my initial answer is consistent with my verification based answer. My initial response states the Chicago Bears won Super Bowl XX, and my revised response also states the Chicago Bears won Super Bowl XX. Therefore, the two responses are consistent with each other. Since my responses are consistent, I am able to answer this question, so I will now provide my answer.

\textbf{$<$/verify$>$} \newline

\textcolor{blue}{The **Chicago Bears** won Super Bowl XX.}     
\end{tcolorbox}

\begin{tcolorbox}[title={\textbf{Example 2:}  $a_0$ is Incorrect and $a_1$ is Inconsistent; In this case the model abstain from answering the question since its revised response is inconsistent with its original response. Model's correct answer is in \textcolor{ForestGreen}{green}, hallucinated responses are in \textcolor{red}{red} while the final model response is in \textcolor{cyan}{blue}.}, colback=red!5!white]

\textbf{Question ($q$):} Who was President when the first Peanuts cartoon was published? 

\textbf{$<$verify$>$} \newline
Let me first try to answer the question. My answer: \textcolor{red}{President Dwight D. Eisenhower was president during the publication of the first Peanuts comic strip in 1950. } \newline

Now, let me try to verify my answer by asking a followup verification question. Given the question, my verification strategy is: **Rephrase Questions** \newline

*Verification Question*: Which U.S. leader held office in the year that the Peanuts comic strip first appeared?

*My response to the Verification Question*: \textcolor{ForestGreen}{**Harry S. Truman** was President of the United States in 1950, when the Peanuts comic strip debuted. } \newline

Now based on the above verification, let me try to answer the question again. My answer based on the verification:
\textcolor{ForestGreen}{Harry S. Truman. } \newline

Now let me check if my initial answer is consistent with my verification based answer. I observe that my initial response names Dwight D. Eisenhower as the president at the time of the first Peanuts cartoon's publication, whereas my revised response states it was Harry S. Truman. Since these are distinct individuals who held the presidency at different times, my two responses are inconsistent. Since my responses are inconsistent, it appears I do not have sufficient knowledge to answer this question, so I am unable to provide an answer. 

\textbf{$<$/verify$>$} \newline

\textcolor{blue}{I do not have sufficient knowledge to answer this question.}

\end{tcolorbox}







\subsection{Data Generation Prompts}

Below are examples of prompts used at different stages of the data generation pipeline.

\begin{tcolorbox}[title={Answer generation}]

Answer the question clearly by providing any relevant and necessary information while avoiding detailed and overly long explanations that are not required to adequately respond to the question. \newline

\{\{Question: ($q$ or $q^{(v)}$)\}\}
\end{tcolorbox}

\begin{tcolorbox}[enhanced, breakable, title=Rephrase strategy verification question $q^{(v)}$ generation for training data augmentation]You are an AI assistant tasked with generating *verification questions* to probe the consistency of a language model's knowledge. The goal is to see if the model gives consistent responses to semantically equivalent but differently phrased questions. \newline

You are given: \newline

* An **original question**. \newline
* The ground truth answer for the question. \newline
* The **answer generated by the model** (which may or may not be correct). \newline

Your task: \newline

1. Explain briefly how you will rephrase the original question to enquire about the same fact from a different angle, without introducing any new information. \newline
2. Then, output a verification question — a reworded version of the original question that explores the same concept or fact, without assuming the model’s answer is correct. The correct answer to this verification question must be the same as for the original question. \newline
3. Ensure the verification question is clear, not overly complex, and does not introduce any new information, new facts, entities, or assumptions that were not already part of the original question. \newline
4. Provide the anwer to the verification question. \newline

While the ground truth answer is provided for reference, the question should not incorporate any information or hints from the answer. \newline

--- \newline
**Example Input:** \newline
**Original Question:** What is the capital of Canada? \newline
**Ground truth Answer:** Ottawa \newline
**Model's Answer:** Toronto is the capital of Canada. \newline

**Example Output:**\newline
**Reasoning:**\newline
Asking about the location of the Parliament and the official seat of government can help verify whether a city is indeed the capital, since these are functions typically associated with a capital city. \newline

**Verification Question:** \verb|`|What Canadian city serves as the official seat of the national government and houses the Parliament buildings\verb|?`| 

**Answer:** Ottawa \newline 

\#\# Original Question: \{\{$q$\}\}

\#\# Ground truth Answer: \{\{Ground truth answer\}\}

\#\# Model's Answer: \{\{$a_0$\}\} \newline

Remember, the verification question should not introduce new information, and it should be answerable by the same ground truth answer as for the original question.
\end{tcolorbox}

\begin{tcolorbox}[title=Revised answer $a_1$ generation]
Answer the question clearly by providing any relevant and necessary information while avoiding detailed and overly long explanations that are not required to adequately respond to the question. \newline

Taking the following verification question and answer into account: \newline
Verification question: \{\{$q^{(v)}$\}\} \newline 
Verification answer: \{\{$a^{(v)}$\}\} \newline

Answer the following question:
\{\{$q$\}\}

\end{tcolorbox}

\begin{tcolorbox}[enhanced, breakable, title=Consistency Judgment ($r$)]
You are a judge model tasked with evaluating the response of an AI assistant after a follow-up verification question. You will be provided with the following: \newline

* The original question. \newline
* The correct answer to the original question, which normally is about an entity, a person, a date, etc.\newline
* The AI assistant’s response to the original question (which may be incorrect). This is called ``Response A".\newline
* The AI assistant’s updated response to the original question, after undergoing a verification step (which may be incorrect). This is called ``Response B".\newline

Your task is to assess the AI assistant’s updated response after the **verification** on two criteria:\newline

1. **Correctness**: Evaluate whether the AI assistant’s updated response after verification is factually correct.\newline
* Use the correct answer to the original question to determine whether the assistant’s updated response after verification matches or aligns with that correct answer.\newline
* Do not reference or take into account the assistant’s response to the original question when making this judgment about the verification answer.\newline

2. **Consistency**: Determine whether the AI assistant’s updated response after verification is consistent with its original response to the original question, i.e. it refers to the same entity, person, date, etc.\newline
* Focus only on whether the two responses present incompatible information. \newline
* Do not consider whether the answers are factually correct or if they match the correct answer—only whether they align or conflict with one another in content or implications.\newline

For each criterion, output your judgment as follows:

\begin{wrapverbatim}
```json
{
  "correctness": {
    "justification": "<Brief explanation>",
    "verdict": "Correct" | "Incorrect"
  },
  "consistency": {
    "justification": "<Brief explanation>",
    "verdict": "Consistent" | "Inconsistent"
  }
}
```
\end{wrapverbatim}

Be concise but clear in your explanations. Base your judgment solely on the provided information.\newline

\#\# Original Question: \{\{$q$\}\} 

\#\# Correct Answer: \{\{Ground truth answer\}\} 

-------

\#\# [Response A]

\#\# Model's Initial Response to the Original Question: \{\{$a_0$\}\}
\newline

\#\# [Response B]

\#\# Model's Updated Response (after verification) to the Original Question: \{\{$a_1$\}\} 

\end{tcolorbox}

\begin{tcolorbox}[enhanced, breakable, title=Answer correctness check for initial response {$a_0$}]
You are a helpful and impartial evaluator tasked with determining whether an AI assistant's response to a given question is factually correct based on a provided ground truth answer. \newline

You will be given: \newline
* A question \newline
* A ground truth answer (i.e., the correct answer) \newline
* The AI assistant’s response \newline

Your task is to: \newline
1. Compare the AI assistant's response to the ground truth answer. \newline
2. Determine if the assistant’s response is factually correct and fully consistent with the ground truth. \newline
3. Judge only factual accuracy — ignore style, formatting, verbosity, or tone. \newline

Return only one of the following three labels: \newline
* Correct — if the assistant’s response is factually consistent with the ground truth. \newline
* Incorrect — if the assistant’s response contains factual errors, omits key information, or contradicts the ground truth. \newline
* Refusal: The response explicitly refrains from answering due to lack of knowledge, uncertainty, or refusal (e.g., “I don’t know,” “I’m not sure,” “I don’t have access to that information”). \newline

Return your response in this format: "\#\# Reason: [your reasoning for the evaluation]; \#\# Label: [your label, Correct or Incorrect or Refusal]" \newline

\#\# [QUESTION]  \newline \{\{ $q$ \}\} \newline

\#\# [GROUND-TRUTH ANSWER] \newline \{\{Ground truth answer\}\} \newline

\#\# [AI ASSISTANT'S RESPONSE] \newline \{\{$a_0$\}\} \newline

\end{tcolorbox}

\begin{tcolorbox}[enhanced, breakable, title=Answer correctness check for verification response {$a^{(v)}$}]
You are an impartial judge. Your task is to evaluate whether an AI Assistant’s response correctly addresses a given question. \newline

You will be given: \newline
* A question \newline
* The AI assistant’s response \newline

Your job is to determine whether the assistant’s response is: \newline
* ``Correct" – The response directly and accurately answers the question. The essential information is correct and relevant, even if phrasing differs or if extra optional context is included. \newline
* ``Incorrect" – The response attempts to answer but contains factual errors, logical errors, incorrect interpretation of the question, or fails to provide the required information. \newline
* ``Refusal" – The response expresses uncertainty, lack of knowledge, or inability to answer (e.g., “I don't know”, “I'm not sure”, “I cannot answer that”, “insufficient information”). \newline

Instructions: \newline
- Base your judgment strictly on the correctness and relevance of the assistant’s answer to the given question. \newline
- Ignore stylistic quality, verbosity, or safety disclaimers unless they interfere with providing the correct answer. \newline
- Provide only one of the labels: Correct, Incorrect, or Refusal. \newline

Return your response in this format: "\#\# Reason: [your reasoning for the evaluation]; \#\# Label: [your label, Correct or Incorrect or Refusal]"\newline

\#\# [QUESTION]  \newline \{\{ $q^{(v)}$ \}\} \newline

\#\# [AI ASSISTANT'S RESPONSE]  \newline \{\{ $a^{(v)}$ \}\} \newline
\end{tcolorbox}

\section{Inference-time Perturbations}
\label{app:shaking_templates}

This section provides the prompt templates used in our \emph{inference-time contextual perturbation} experiments. Contextual perturbations are applied solely at inference time and do not involve any modification to model parameters or decoding hyperparameters. Instead, they alter the generation context in a controlled manner to induce drift in the model’s inference trajectory.
To induce perturbation, we modify the chat template by removing the \texttt{end\_of\_turn} marker from the assistant response, causing the model to continue generation from previously produced text. 
We evaluate two perturbation variants:
\begin{itemize}
    \item The first applies contextual perturbation immediately prior to answering the user query (\emph{Pert}).
    
    \item The second introduces a stronger intervention by including an explicit draft response $a_0$ to the original query before applying perturbation (\emph{$a_0$+Pert}). This allows us to assess whether intervening after a provisional answer has been formed alters the effect of inference-time perturbations.
\end{itemize}   In all cases, the auxiliary content is not visible in the final response.
Below are the prompt templates used when running the \emph{inference shaking} ablations.

\begin{tcolorbox}[enhanced, breakable, title=Perturbation Only, colback=yellow!10!white]
Let's first answer another question. \newline \newline
*Question*: \{\{$q'$\}\} \newline
*Answer*: \{\{$a'$\}\} \newline 

Now let's answer the user's question.

\end{tcolorbox}

\begin{tcolorbox}[enhanced, breakable, title=Draft Response $+$ Perturbation, colback=yellow!10!white]
Let's first generate a draft response: \{\{$a_0$\}\} \newline

Now let's answer another question. \newline

*Question*: \{\{$q'$\}\} \newline
*Answer*: \{\{$a'$\}\} \newline 

Now let's generate a final answer to the user's question.
\end{tcolorbox}



\section{Baselines}
\label{sec:appendix_baselines}

\subsection{Knowledge probing}
Knowledge Probing (KP)~\cite{grattafiori2024llama} is a training-time approach that uses model-generated responses as a starting point. First, the model’s responses to a set of prompts are generated and annotated as correct or incorrect. Then, the model is fine-tuned on these examples: it is trained to produce the correct answer when one exists and to refrain from responding when the answer is incorrect or unknown. Algorithm~\ref{alg:knowledge_probing} illustrates this process.

\begin{algorithm}[H]
    \caption{Knowledge Probing Training Procedure}
    \label{alg:knowledge_probing}
    \begin{algorithmic}[1]
        \REQUIRE Factoid QA dataset $\mathcal{D} = \{(q, y)\}$, base model $M$
        \FOR{each question $q$ with reference answer $y$ in $\mathcal{D}$}
            \STATE Sample the model’s response $a \sim M(q)$
            \STATE Evaluate correctness of $a$ against the reference answer $y$
            \IF{$a$ is incorrect}
                \STATE Replace $a$ with a refusal response (e.g., ``I don't know.'')
            \ENDIF
            \STATE Add $(q, a)$ to the supervised fine-tuning dataset
        \ENDFOR
        \STATE Fine-tune $M$ on the constructed dataset
    \end{algorithmic}
\end{algorithm}

\subsection{Self-Consistency}
Self-Consistency (\systemSC{}-$k$) is a post-hoc approach that aims to improve the reliability of model outputs without modifying the model’s training. For each prompt, the model first generates a single response at temperature $0.0$, and then $k$ additional responses are sampled at temperature $1.0$. These $k$ responses are compared to the temperature $0.0$ response using the SelfCheckGPT consistency check~\cite{manakul2023selfcheckgpt}, which evaluates whether the factoids in the responses agree. If at least half of the $k$ generations support the answer, it is considered non-hallucinated and returned; otherwise, the model abstains. The system prompt used for consistency checking is provided below.

\begin{tcolorbox}[enhanced, breakable, title=Self-Consistency Check, colback=yellow!10!white]
You are an expert QA judge. Your job is to determine whether two responses to the same question are consistent at the level of the answer entity.\newline\newline

The answer entity is the entity that directly answers the question being asked (e.g., the person, group, organization, or place requested by the question). Your evaluation must be based only on whether both responses identify the same answer entity.\newline\newline
You will be given: \newline
* A question \newline
* A reference response \newline
* A candidate response \newline

The answer entity is the entity that directly answers the question being asked (e.g., the person, group, organization, or place requested by the question). \newline

Your job is to determine whether the candidate response is consistent with the reference response with respect to the answer entity. \newline

Ignore factual correctness entirely. Ignore numbers, dates, counts, causes, outcomes, and details. Ignore uncertainty, premise rejection, explanations, tone, formatting, and emojis. \newline

Your response must be exactly one of the following labels: \newline
* ``CONSISTENT" \newline
* ``INCONSISTENT" \newline
* ``NOT\_ATTEMPTED" \newline

Label definitions: \newline
* ``CONSISTENT" – Both responses identify or clearly point to the same answer entity that the question asks for, even if they disagree on facts, details, or whether the event occurred. \newline
* ``INCONSISTENT" – The responses identify different answer entities for the question. \newline
* ``NOT\_ATTEMPTED" – The candidate response provides no identifiable answer entity to the question (e.g., refusal, empty output, or irrelevant content). Do \textbf{not} consider the reference response when assigning NOT\_ATTEMPTED. \newline

Return your response in this format: "\#\# Reason: [your reasoning]; \#\# Label: [CONSISTENT or INCONSISTENT or NOT\_ATTEMPTED]" \newline

\#\# [QUESTION] \newline
\{\{ $q^{(v)}$ \}\} \newline

\#\# [REFERENCE RESPONSE] \newline
\{\{ $a^{(0)}$ \}\} \newline

\#\# [CANDIDATE RESPONSE] \newline
\{\{ $a^{(k)}$ \}\}
\end{tcolorbox}

\section{Training and Optimization Details}
\label{sec:appendix_training_parameters}

All models are trained using full-parameter supervised fine-tuning (SFT) with bfloat16 precision and a maximum sequence length of 3{,}000 tokens. Optimization is performed using AdamW with an initial learning rate of $5\times10^{-7}$, cosine decay scheduling with a minimum learning rate of $5\times10^{-8}$, and a warmup ratio of 0.1. Models are trained for two epochs with a fixed effective batch size of 640 across all experiments. Distributed training is enabled via DeepSpeed ZeRO-3, and FlashAttention-2 is employed to improve memory efficiency and throughput. Model-specific instruction and loss templates are used consistently for each backbone.

Our study includes 8 backbone models, each evaluated under 4 training strategies, resulting in 32 full fine-tuning runs. Training is conducted on NVIDIA H200 GPUs (approximately 140\,GB memory per device) using multi-GPU distributed setups. Each run requires roughly 12 hours on 4 GPUs, and the complete experimental suite consumed over 2{,}000 GPU-hours in total. This compute budget reflects the cost of full-parameter optimization across multiple model families and training configurations, rather than parameter-efficient adaptation.

\section{Extended Experimental Results}
\label{sec:appendix_generalization_results}
Across all out-of-domain benchmarks and model families, the extended results in Table~\ref{tab:evaluation_triviaqa_appendix}-\ref{tab:evaluation_freshqa_2025_11_24_appendix} reinforce the main findings: \system{} consistently improves the precision–recall trade-off relative to both base models, inference-only baselines, and abstention-based methods. Also, we observe a consistent pattern in the interaction between coverage and selective accuracy (precision). Coverage is defined as the proportion of questions that were attempted, which, together with selective accuracy, are usually used in risk-coverage curves~\cite{JMLR:v11:el-yaniv10a}. \[
   \mathcal{C}ov = 1 - \frac{\#R}{\#(C+X+R)}
\]

High coverage obtained by always answering leads to excessive hallucinations, while overly conservative refusal strategies collapse recall and F1. Base models consistently exhibit near-maximal coverage ($>90\%$ in most settings) but low precision, particularly on HotpotQA and NQ-Open, indicating a strong tendency to answer out-of-domain questions without reliable knowledge. Knowledge probing substantially improves precision but does so by sharply reducing coverage—often below 30\% on HotpotQA, resulting in degraded F1 despite high $R^{+}$. In contrast, \system{} variants achieve a more favorable balance. Across all datasets, structured masking (\systemSM{}, \systemMM{}) consistently reduces incorrect responses ($X$) and hallucinated acceptances ($R^{-}$) while preserving substantially higher coverage than other baselines.

\begin{table*}[t]
\centering
\caption{Evaluation on the Triviaqa benchmark across small (2B--4B) and mid-sized (7B--12B) instruction-tuned models.}
\label{tab:evaluation_triviaqa_appendix}
\footnotesize
\setlength{\tabcolsep}{2.5pt}
\renewcommand{\arraystretch}{0.95}
\begin{tabular}{l r|r|rr r r c|m{1.1cm} | l r|r|rr r r c|c}
\toprule
\multicolumn{9}{c}{\textbf{Small-Size Models (2B-4B)}} & \multicolumn{9}{c}{\textbf{Mid-Size Models (7B-12B)}} \\
\cmidrule(lr){1-9} \cmidrule(lr){10-18}
\textbf{Model / Setup} & $C${\tiny$\uparrow$} & $X${\tiny$\downarrow$} & $R^{-}${\tiny$\downarrow$} & $R^{+}${\tiny$\uparrow$} & $\mathcal{C}ov$ & $\mathcal{P}${\tiny$\uparrow$} & $\mathcal{R}${\tiny$\uparrow$} & F1{\tiny$\uparrow$} & \textbf{Model / Setup} & $C${\tiny$\uparrow$} & $X${\tiny$\downarrow$} & $R^{-}${\tiny$\downarrow$} & $R^{+}${\tiny$\uparrow$} & $\mathcal{C}ov$ & $\mathcal{P}${\tiny$\uparrow$} & $\mathcal{R}${\tiny$\uparrow$} & F1{\tiny$\uparrow$} \\
\midrule
\textbf{Gemma-2-2b} &  &  &  &  &  &  &  &  & \textbf{Gemma-2-9b} &  &  &  &  &  &  &  &  \\
\rowcolor[gray]{0.85}\quad \textit{Base} & \textit{58.5} & \textit{40.8} & \multicolumn{2}{c}{(\textit{0.7})} & \textit{99.3} & \textit{58.9} & - & - & \quad \textit{Base} & \textit{77.0} & \textit{22.0} & \multicolumn{2}{c}{(\textit{0.9})} & \textit{99.1} & \textit{77.8} & - & - \\
\quad \systemKP{} & 46.6 & \textbf{13.2} & 12.6 & \textbf{27.7} & 59.8 & 78.0 & 78.8 & 78.4 & \quad \systemKP{} & 64.3 & 6.9 & 14.0 & \textbf{14.7} & 71.2 & 90.3 & 82.1 & 86.0 \\
\quad \systemNM{} & 49.3 & \textbf{13.2} & 10.2 & 27.4 & 62.4 & \textbf{78.9} & 82.9 & 80.8 & \quad \systemNM{} & 70.5 & \textbf{6.6} & 8.5 & 14.4 & 77.1 & \textbf{91.5} & 89.2 & 90.3 \\
\quad \systemMM{} & 54.4 & 17.5 & 6.3 & 21.8 & 71.9 & 75.6 & 89.7 & \textbf{82.0} & \quad \systemMM{} & 74.3 & 9.7 & 5.1 & 10.9 & 84.0 & 88.4 & 93.5 & 90.9 \\
\quad \systemSM{} & \textbf{55.5} & 19.6 & \textbf{5.4} & 19.6 & \textbf{75.0} & 73.9 & \textbf{91.2} & 81.6 & \quad \systemSM{} & \textbf{74.7} & 10.0 & \textbf{4.7} & 10.6 & \textbf{84.7} & 88.2 & \textbf{94.1} & \textbf{91.1} \\
\midrule
\textbf{Qwen2.5-3B} &  &  &  &  &  &  &  &  & \textbf{Qwen2.5-7B} &  &  &  &  &  &  &  &  \\
\rowcolor[gray]{0.85}\quad \textit{Base} & \textit{55.5} & \textit{40.5} & \multicolumn{2}{c}{(\textit{4.0})} & \textit{96.0} & \textit{57.8} & - & - & \quad \textit{Base} & \textit{67.1} & \textit{32.2} & \multicolumn{2}{c}{(\textit{0.8})} & \textit{99.2} & \textit{67.6} & - & - \\
\quad \systemKP{} & 33.5 & \textbf{5.7} & 22.3 & \textbf{38.5} & 39.2 & \textbf{85.5} & 60.0 & 70.5 & \quad \systemKP{} & 52.9 & \textbf{7.9} & 14.1 & \textbf{25.0} & 60.8 & \textbf{87.0} & 78.9 & 82.8 \\
\quad \systemNM{} & 46.6 & 11.0 & 10.4 & 32.0 & 57.6 & 80.9 & 81.7 & 81.3 & \quad \systemNM{} & 59.3 & 8.9 & 8.8 & 23.0 & 68.2 & 86.9 & 87.1 & 87.0 \\
\quad \systemMM{} & 50.0 & 13.7 & 8.3 & 28.0 & 63.7 & 78.5 & 85.7 & \textbf{82.0} & \quad \systemMM{} & 60.9 & 9.8 & 7.6 & 21.7 & 70.8 & 86.1 & 88.9 & \textbf{87.5} \\
\quad \systemSM{} & \textbf{52.2} & 19.1 & \textbf{5.8} & 22.9 & \textbf{71.3} & 73.2 & \textbf{90.0} & 80.7 & \quad \systemSM{} & \textbf{61.1} & 10.5 & \textbf{7.3} & 21.0 & \textbf{71.6} & 85.3 & \textbf{89.3} & 87.3 \\
\midrule
\textbf{Llama-3.2-3B} &  &  &  &  &  &  &  &  & \textbf{Llama-3.1-8B} &  &  &  &  &  &  &  &  \\
\rowcolor[gray]{0.85}\quad \textit{Base} & \textit{63.5} & \textit{21.4} & \multicolumn{2}{c}{(\textit{15.2})} & \textit{84.8} & \textit{74.8} & - & - & \quad \textit{Base} & \textit{79.9} & \textit{14.8} & \multicolumn{2}{c}{(\textit{5.3})} & \textit{94.7} & \textit{84.4} & - & - \\
\quad \systemKP{} & 54.3 & \textbf{11.4} & 10.1 & \textbf{24.2} & 65.7 & 82.7 & 84.3 & 83.5 & \quad \systemKP{} & 70.2 & \textbf{9.5} & 8.8 & \textbf{11.5} & 79.7 & 88.1 & 88.9 & 88.5 \\
\quad \systemNM{} & 60.5 & 14.9 & \textbf{6.1} & 18.5 & \textbf{75.3} & 80.2 & 90.8 & 85.2 & \quad \systemNM{} & 73.8 & 10.4 & 6.7 & 9.1 & 84.2 & 87.6 & 91.7 & 89.6 \\
\quad \systemMM{} & 62.3 & 12.4 & 6.3 & 18.9 & 74.7 & \textbf{83.4} & 90.8 & \textbf{86.9} & \quad \systemMM{} & 75.0 & 10.1 & 6.1 & 8.9 & 85.1 & 88.2 & 92.5 & 90.3 \\
\quad \systemSM{} & \textbf{62.4} & 12.7 & 6.3 & 18.6 & 75.1 & 83.1 & \textbf{90.9} & 86.8 & \quad \systemSM{} & \textbf{75.8} & 9.7 & \textbf{5.3} & 9.2 & \textbf{85.5} & \textbf{88.7} & \textbf{93.5} & \textbf{91.0} \\
\midrule
\textbf{Gemma-3-4b} &  &  &  &  &  &  &  &  & \textbf{Gemma-3-12b} &  &  &  &  &  &  &  &  \\
\rowcolor[gray]{0.85}\quad \textit{Base} & \textit{57.1} & \textit{42.8} & \multicolumn{2}{c}{(\textit{0.1})} & \textit{99.9} & \textit{57.1} & - & - & \quad \textit{Base} & \textit{76.8} & \textit{23.1} & \multicolumn{2}{c}{(\textit{0.1})} & \textit{99.9} & \textit{76.9} & - & - \\
\quad \systemKP{} & 41.1 & 8.1 & 16.4 & 34.4 & 49.3 & 83.5 & 71.6 & 77.1 & \quad \systemKP{} & 62.4 & \textbf{6.4} & 14.5 & \textbf{16.8} & 68.7 & 90.7 & 81.2 & 85.7 \\
\quad \systemNM{} & 44.1 & \textbf{6.7} & 14.5 & \textbf{34.7} & 50.8 & \textbf{86.8} & 75.2 & 80.6 & \quad \systemNM{} & 70.8 & 10.1 & 6.5 & 12.6 & \textbf{80.9} & 87.5 & 91.5 & 89.5 \\
\quad \systemMM{} & \textbf{52.5} & 14.1 & \textbf{8.3} & 25.0 & \textbf{66.6} & 78.8 & \textbf{86.3} & 82.4 & \quad \systemMM{} & \textbf{72.0} & 8.8 & \textbf{6.4} & 12.8 & 80.7 & 89.1 & \textbf{91.8} & 90.5 \\
\quad \systemSM{} & 51.7 & 10.4 & 9.2 & 28.7 & 62.1 & 83.2 & 85.0 & \textbf{84.1} & \quad \systemSM{} & 70.9 & 7.2 & 7.7 & 14.2 & 78.0 & \textbf{90.8} & 90.2 & \textbf{90.5} \\
\bottomrule
\end{tabular}
\end{table*}

\begin{table*}[t]
\centering
\caption{Generalization on the Hotpotqa benchmark across small (2B--4B) and mid-sized (7B--12B) instruction-tuned models.}
\label{tab:evaluation_hotpotqa_appendix}
\footnotesize
\setlength{\tabcolsep}{2.5pt}
\renewcommand{\arraystretch}{0.95}
\begin{tabular}{l r|r|rr r r c|m{1.1cm} | l r|r|rr r r c|c}
\toprule
\multicolumn{9}{c}{\textbf{Small-Size Models (2B-4B)}} & \multicolumn{9}{c}{\textbf{Mid-Size Models (7B-12B)}} \\
\cmidrule(lr){1-9} \cmidrule(lr){10-18}
\textbf{Model / Setup} & $C${\tiny$\uparrow$} & $X${\tiny$\downarrow$} & $R^{-}${\tiny$\downarrow$} & $R^{+}${\tiny$\uparrow$} & $\mathcal{C}ov$ & $\mathcal{P}${\tiny$\uparrow$} & $\mathcal{R}${\tiny$\uparrow$} & F1{\tiny$\uparrow$} & \textbf{Model / Setup} & $C${\tiny$\uparrow$} & $X${\tiny$\downarrow$} & $R^{-}${\tiny$\downarrow$} & $R^{+}${\tiny$\uparrow$} & $\mathcal{C}ov$ & $\mathcal{P}${\tiny$\uparrow$} & $\mathcal{R}${\tiny$\uparrow$} & F1{\tiny$\uparrow$} \\
\midrule
\textbf{Gemma-2-2b} &  &  &  &  &  &  &  &  & \textbf{Gemma-2-9b} &  &  &  &  &  &  &  &  \\
\rowcolor[gray]{0.85}\quad \textit{Base} & \textit{28.1} & \textit{61.3} & \multicolumn{2}{c}{(\textit{10.6})} & \textit{89.4} & \textit{31.4} & - & - & \quad \textit{Base} & \textit{37.0} & \textit{46.7} & \multicolumn{2}{c}{(\textit{16.3})} & \textit{83.7} & \textit{44.2} & - & - \\
\quad \systemKP{} & 12.0 & \textbf{11.2} & 15.8 & \textbf{61.0} & 23.1 & \textbf{51.7} & 43.0 & 47.0 & \quad \systemKP{} & 15.7 & \textbf{8.5} & 21.2 & \textbf{54.6} & 24.1 & \textbf{64.9} & 42.5 & 51.3 \\
\quad \systemNM{} & 17.0 & 17.8 & 11.4 & 53.9 & 34.8 & 48.8 & 59.9 & 53.8 & \quad \systemNM{} & 26.1 & 16.1 & 11.4 & 46.4 & 42.2 & 61.9 & 69.7 & 65.5 \\
\quad \systemMM{} & 20.9 & 28.4 & 7.5 & 43.2 & 49.2 & 42.4 & 73.6 & \textbf{53.8} & \quad \systemMM{} & 30.2 & 23.7 & 7.2 & 38.9 & 53.9 & 56.0 & 80.7 & \textbf{66.1} \\
\quad \systemSM{} & \textbf{22.1} & 32.6 & \textbf{6.3} & 39.0 & \textbf{54.7} & 40.4 & \textbf{77.9} & 53.2 & \quad \systemSM{} & \textbf{30.8} & 25.9 & \textbf{6.8} & 36.5 & \textbf{56.6} & 54.3 & \textbf{81.9} & 65.3 \\
\midrule
\textbf{Qwen2.5-3B} &  &  &  &  &  &  &  &  & \textbf{Qwen2.5-7B} &  &  &  &  &  &  &  &  \\
\rowcolor[gray]{0.85}\quad \textit{Base} & \textit{22.3} & \textit{53.5} & \multicolumn{2}{c}{(\textit{24.2})} & \textit{75.8} & \textit{29.4} & - & - & \quad \textit{Base} & \textit{31.5} & \textit{58.9} & \multicolumn{2}{c}{(\textit{9.6})} & \textit{90.4} & \textit{34.9} & - & - \\
\quad \systemKP{} & 6.5 & \textbf{3.6} & 16.2 & \textbf{73.7} & 10.1 & \textbf{64.6} & 28.8 & 39.8 & \quad \systemKP{} & 14.0 & \textbf{8.7} & 17.2 & \textbf{60.2} & 22.6 & \textbf{61.7} & 44.8 & 51.9 \\
\quad \systemNM{} & 13.6 & 12.6 & 10.0 & 63.8 & 26.2 & 51.9 & 57.6 & \textbf{54.6} & \quad \systemNM{} & 21.4 & 15.7 & 10.7 & 52.3 & 37.1 & 57.7 & 66.8 & 61.9 \\
\quad \systemMM{} & 17.3 & 23.3 & 7.2 & 52.2 & 40.6 & 42.6 & 70.6 & 53.1 & \quad \systemMM{} & 23.5 & 20.3 & 8.9 & 47.3 & 43.7 & 53.6 & 72.4 & 61.6 \\
\quad \systemSM{} & \textbf{19.5} & 31.8 & \textbf{5.0} & 43.8 & \textbf{51.2} & 38.0 & \textbf{79.5} & 51.4 & \quad \systemSM{} & \textbf{24.8} & 21.8 & \textbf{8.1} & 45.3 & \textbf{46.6} & 53.2 & \textbf{75.3} & \textbf{62.4} \\
\midrule
\textbf{Llama-3.2-3B} &  &  &  &  &  &  &  &  & \textbf{Llama-3.1-8B} &  &  &  &  &  &  &  &  \\
\rowcolor[gray]{0.85}\quad \textit{Base} & \textit{21.4} & \textit{35.0} & \multicolumn{2}{c}{(\textit{43.5})} & \textit{56.5} & \textit{38.0} & - & - & \quad \textit{Base} & \textit{32.2} & \textit{32.4} & \multicolumn{2}{c}{(\textit{35.4})} & \textit{64.7} & \textit{49.9} & - & - \\
\quad \systemKP{} & 13.2 & \textbf{15.6} & 9.3 & \textbf{61.9} & 28.8 & 45.8 & 58.6 & 51.4 & \quad \systemKP{} & 22.0 & \textbf{18.3} & 10.8 & \textbf{49.0} & 40.2 & \textbf{54.6} & 67.1 & 60.2 \\
\quad \systemNM{} & 17.4 & 19.9 & 6.4 & 56.3 & 37.2 & 46.7 & 73.0 & 56.9 & \quad \systemNM{} & 24.7 & 23.8 & 9.1 & 42.4 & 48.4 & 50.9 & 73.0 & 60.0 \\
\quad \systemMM{} & \textbf{18.9} & 20.1 & 6.4 & 54.7 & 39.0 & \textbf{48.4} & 74.8 & \textbf{58.8} & \quad \systemMM{} & \textbf{30.4} & 30.2 & 6.0 & 33.4 & \textbf{60.6} & 50.2 & \textbf{83.4} & \textbf{62.7} \\
\quad \systemSM{} & 18.7 & 23.0 & \textbf{6.1} & 52.2 & \textbf{41.7} & 44.9 & \textbf{75.3} & 56.3 & \quad \systemSM{} & 28.1 & 28.9 & \textbf{5.8} & 37.1 & 57.0 & 49.2 & 82.8 & 61.7 \\
\midrule
\textbf{Gemma-3-4b} &  &  &  &  &  &  &  &  & \textbf{Gemma-3-12b} &  &  &  &  &  &  &  &  \\
\rowcolor[gray]{0.85}\quad \textit{Base} & \textit{27.1} & \textit{72.3} & \multicolumn{2}{c}{(\textit{0.5})} & \textit{99.5} & \textit{27.3} & - & - & \quad \textit{Base} & \textit{37.6} & \textit{62.1} & \multicolumn{2}{c}{(\textit{0.2})} & \textit{99.8} & \textit{37.7} & - & - \\
\quad \systemKP{} & 10.2 & \textbf{9.6} & 16.6 & \textbf{63.5} & 19.9 & 51.5 & 38.1 & 43.8 & \quad \systemKP{} & 19.4 & \textbf{11.0} & 17.9 & \textbf{51.7} & 30.4 & \textbf{63.9} & 52.0 & 57.4 \\
\quad \systemNM{} & 14.6 & 12.4 & 12.5 & 60.5 & 27.0 & \textbf{54.1} & 54.0 & 54.1 & \quad \systemNM{} & 28.7 & 26.4 & 8.8 & 36.1 & \textbf{55.0} & 52.1 & 76.4 & 62.0 \\
\quad \systemMM{} & \textbf{18.9} & 25.0 & 9.2 & 46.9 & \textbf{43.9} & 43.1 & 67.2 & 52.5 & \quad \systemMM{} & \textbf{29.7} & 25.3 & \textbf{8.6} & 36.4 & 55.0 & 54.0 & \textbf{77.5} & 63.6 \\
\quad \systemSM{} & 18.8 & 20.9 & \textbf{9.1} & 51.1 & 39.7 & 47.3 & \textbf{67.3} & \textbf{55.5} & \quad \systemSM{} & 29.0 & 21.7 & 9.4 & 40.0 & 50.6 & 57.2 & 75.6 & \textbf{65.1} \\
\bottomrule
\end{tabular}
\end{table*}

\begin{table*}[t]
\centering
\caption{Generalization on the Natural Questions benchmark across small (2B--4B) and mid-sized (7B--12B) instruction-tuned models.}
\label{tab:evaluation_nqopen_appendix}
\footnotesize
\setlength{\tabcolsep}{2.5pt}
\renewcommand{\arraystretch}{0.95}
\begin{tabular}{l r|r|rr r r c|m{1.1cm} | l r|r|rr r r c|c}
\toprule
\multicolumn{9}{c}{\textbf{Small-Size Models (2B-4B)}} & \multicolumn{9}{c}{\textbf{Mid-Size Models (7B-12B)}} \\
\cmidrule(lr){1-9} \cmidrule(lr){10-18}
\textbf{Model / Setup} & $C${\tiny$\uparrow$} & $X${\tiny$\downarrow$} & $R^{-}${\tiny$\downarrow$} & $R^{+}${\tiny$\uparrow$} & $\mathcal{C}ov$ & $\mathcal{P}${\tiny$\uparrow$} & $\mathcal{R}${\tiny$\uparrow$} & F1{\tiny$\uparrow$} & \textbf{Model / Setup} & $C${\tiny$\uparrow$} & $X${\tiny$\downarrow$} & $R^{-}${\tiny$\downarrow$} & $R^{+}${\tiny$\uparrow$} & $\mathcal{C}ov$ & $\mathcal{P}${\tiny$\uparrow$} & $\mathcal{R}${\tiny$\uparrow$} & F1{\tiny$\uparrow$} \\
\midrule
\textbf{Gemma-2-2b} &  &  &  &  &  &  &  &  & \textbf{Gemma-2-9b} &  &  &  &  &  &  &  &  \\
\rowcolor[gray]{0.85}\quad \textit{Base} & \textit{38.0} & \textit{56.0} & \multicolumn{2}{c}{(\textit{6.1})} & \textit{93.9} & \textit{40.4} & - & - & \quad \textit{Base} & \textit{49.7} & \textit{43.2} & \multicolumn{2}{c}{(\textit{7.1})} & \textit{92.9} & \textit{53.5} & - & - \\
\quad \systemKP{} & 21.6 & \textbf{16.2} & 15.9 & \textbf{46.4} & 37.8 & \textbf{57.1} & 57.6 & 57.3 & \quad \systemKP{} & 30.1 & \textbf{12.6} & 19.9 & \textbf{37.5} & 42.7 & \textbf{70.5} & 60.2 & 64.9 \\
\quad \systemNM{} & 28.1 & 25.6 & 9.4 & 36.9 & 53.7 & 52.3 & 74.9 & 61.6 & \quad \systemNM{} & 41.7 & 20.9 & 8.6 & 28.8 & 62.6 & 66.6 & 82.9 & 73.8 \\
\quad \systemMM{} & 30.5 & 30.9 & 7.2 & 31.4 & 61.4 & 49.6 & 80.8 & 61.5 & \quad \systemMM{} & 43.7 & 24.3 & 6.6 & 25.5 & 67.9 & 64.3 & 86.9 & \textbf{73.9} \\
\quad \systemSM{} & \textbf{32.2} & 34.0 & \textbf{5.2} & 28.6 & \textbf{66.2} & 48.7 & \textbf{86.0} & \textbf{62.2} & \quad \systemSM{} & \textbf{44.0} & 26.7 & \textbf{5.6} & 23.7 & \textbf{70.8} & 62.2 & \textbf{88.8} & 73.2 \\
\midrule
\textbf{Qwen2.5-3B} &  &  &  &  &  &  &  &  & \textbf{Qwen2.5-7B} &  &  &  &  &  &  &  &  \\
\rowcolor[gray]{0.85}\quad \textit{Base} & \textit{34.0} & \textit{54.9} & \multicolumn{2}{c}{(\textit{11.1})} & \textit{88.9} & \textit{38.2} & - & - & \quad \textit{Base} & \textit{42.6} & \textit{55.1} & \multicolumn{2}{c}{(\textit{2.2})} & \textit{97.8} & \textit{43.6} & - & - \\
\quad \systemKP{} & 19.0 & \textbf{12.3} & 13.7 & \textbf{55.0} & 31.3 & \textbf{60.7} & 58.0 & 59.3 & \quad \systemKP{} & 27.6 & \textbf{14.0} & 14.1 & \textbf{44.3} & 41.6 & \textbf{66.3} & 66.2 & 66.3 \\
\quad \systemNM{} & 24.5 & 19.4 & 9.3 & 46.8 & 43.9 & 55.8 & 72.5 & 63.1 & \quad \systemNM{} & 34.0 & 19.8 & 9.0 & 37.2 & 53.9 & 63.2 & 79.1 & 70.3 \\
\quad \systemMM{} & 26.6 & 22.5 & 7.8 & 43.2 & 49.0 & 54.2 & 77.4 & \textbf{63.7} & \quad \systemMM{} & 34.2 & 21.7 & 9.1 & 35.0 & 55.9 & 61.2 & 79.0 & 69.0 \\
\quad \systemSM{} & \textbf{28.6} & 31.5 & \textbf{5.1} & 34.8 & \textbf{60.1} & 47.5 & \textbf{84.8} & 60.9 & \quad \systemSM{} & \textbf{36.1} & 22.2 & \textbf{7.3} & 34.4 & \textbf{58.4} & 61.9 & \textbf{83.2} & \textbf{71.0} \\
\midrule
\textbf{Llama-3.2-3B} &  &  &  &  &  &  &  &  & \textbf{Llama-3.1-8B} &  &  &  &  &  &  &  &  \\
\rowcolor[gray]{0.85}\quad \textit{Base} & \textit{58.2} & \textit{37.9} & \multicolumn{2}{c}{(\textit{4.0})} & \textit{96.0} & \textit{60.6} & - & - & \quad \textit{Base} & \textit{64.7} & \textit{33.2} & \multicolumn{2}{c}{(\textit{2.1})} & \textit{97.9} & \textit{66.0} & - & - \\
\quad \systemKP{} & 40.6 & \textbf{20.2} & 15.4 & \textbf{23.8} & 60.8 & \textbf{66.8} & 72.5 & 69.5 & \quad \systemKP{} & 48.0 & \textbf{19.7} & 14.2 & \textbf{18.1} & 67.7 & \textbf{70.9} & 77.1 & 73.9 \\
\quad \systemNM{} & 49.1 & 27.4 & 7.8 & 15.7 & 76.5 & 64.1 & 86.3 & 73.6 & \quad \systemNM{} & \textbf{58.6} & 29.1 & \textbf{4.9} & 7.5 & \textbf{87.6} & 66.8 & \textbf{92.3} & 77.5 \\
\quad \systemMM{} & 49.7 & 26.2 & 8.1 & 15.9 & 76.0 & 65.5 & 86.0 & 74.3 & \quad \systemMM{} & 56.6 & 27.3 & 7.3 & 8.9 & 83.9 & 67.5 & 88.6 & 76.6 \\
\quad \systemSM{} & \textbf{50.1} & 26.7 & \textbf{7.5} & 15.7 & \textbf{76.8} & 65.3 & \textbf{87.0} & \textbf{74.6} & \quad \systemSM{} & 57.6 & 27.1 & 6.2 & 9.1 & 84.7 & 68.1 & 90.3 & \textbf{77.6} \\
\midrule
\textbf{Gemma-3-4b} &  &  &  &  &  &  &  &  & \textbf{Gemma-3-12b} &  &  &  &  &  &  &  &  \\
\rowcolor[gray]{0.85}\quad \textit{Base} & \textit{36.3} & \textit{63.5} & \multicolumn{2}{c}{(\textit{0.2})} & \textit{99.8} & \textit{36.3} & - & - & \quad \textit{Base} & \textit{50.1} & \textit{49.8} & \multicolumn{2}{c}{(\textit{0.1})} & \textit{99.9} & \textit{50.1} & - & - \\
\quad \systemKP{} & 20.2 & 15.8 & 15.5 & 48.4 & 36.1 & 56.1 & 56.5 & 56.3 & \quad \systemKP{} & 33.5 & \textbf{16.4} & 16.1 & \textbf{34.1} & 49.9 & \textbf{67.1} & 67.6 & 67.3 \\
\quad \systemNM{} & 24.2 & \textbf{15.2} & 12.0 & \textbf{48.7} & 39.3 & \textbf{61.5} & 66.9 & 64.0 & \quad \systemNM{} & 40.8 & 24.9 & 8.9 & 25.5 & \textbf{65.7} & 62.1 & 82.2 & 70.8 \\
\quad \systemMM{} & 26.4 & 23.1 & 10.5 & 40.0 & \textbf{49.5} & 53.3 & 71.5 & 61.1 & \quad \systemMM{} & \textbf{42.0} & 23.1 & \textbf{8.2} & 26.7 & 65.1 & 64.5 & \textbf{83.7} & 72.8 \\
\quad \systemSM{} & \textbf{26.8} & 19.5 & \textbf{9.9} & 43.7 & 46.3 & 57.9 & \textbf{73.0} & \textbf{64.6} & \quad \systemSM{} & 41.3 & 21.0 & 9.3 & 28.4 & 62.3 & 66.3 & 81.6 & \textbf{73.2} \\
\bottomrule
\end{tabular}
\end{table*}

\begin{table*}[t]
\centering
\caption{Generalization on the FreshQA (2025/11/24) benchmark across small (2B--4B) and mid-sized (7B--12B) instruction-tuned models.}
\label{tab:evaluation_freshqa_2025_11_24_appendix}
\footnotesize
\setlength{\tabcolsep}{2.5pt}
\renewcommand{\arraystretch}{0.95}
\begin{tabular}{l r|r|rr r r c|m{1.1cm} | l r|r|rr r r c|c}
\toprule
\multicolumn{9}{c}{\textbf{Small-Size Models (2B-4B)}} & \multicolumn{9}{c}{\textbf{Mid-Size Models (7B-12B)}} \\
\cmidrule(lr){1-9} \cmidrule(lr){10-18}
\textbf{Model / Setup} & $C${\tiny$\uparrow$} & $X${\tiny$\downarrow$} & $R^{-}${\tiny$\downarrow$} & $R^{+}${\tiny$\uparrow$} & $\mathcal{C}ov$ & $\mathcal{P}${\tiny$\uparrow$} & $\mathcal{R}${\tiny$\uparrow$} & F1{\tiny$\uparrow$} & \textbf{Model / Setup} & $C${\tiny$\uparrow$} & $X${\tiny$\downarrow$} & $R^{-}${\tiny$\downarrow$} & $R^{+}${\tiny$\uparrow$} & $\mathcal{C}ov$ & $\mathcal{P}${\tiny$\uparrow$} & $\mathcal{R}${\tiny$\uparrow$} & F1{\tiny$\uparrow$} \\
\midrule
\textbf{Gemma-2-2b} &  &  &  &  &  &  &  &  & \textbf{Gemma-2-9b} &  &  &  &  &  &  &  &  \\
\rowcolor[gray]{0.85}\quad \textit{Base} & \textit{34.6} & \textit{58.2} & \multicolumn{2}{c}{(\textit{7.2})} & \textit{92.8} & \textit{37.3} & - & - & \quad \textit{Base} & \textit{44.0} & \textit{44.0} & \multicolumn{2}{c}{(\textit{12.0})} & \textit{88.0} & \textit{50.0} & - & - \\
\quad \systemKP{} & 23.6 & \textbf{20.0} & 10.4 & \textbf{46.0} & 43.6 & \textbf{54.1} & 69.4 & 60.8 & \quad \systemKP{} & 26.2 & \textbf{12.8} & 16.6 & \textbf{44.4} & 39.0 & \textbf{67.2} & 61.2 & 64.1 \\
\quad \systemNM{} & 23.6 & 27.6 & 10.8 & 38.0 & 51.2 & 46.1 & 68.6 & 55.1 & \quad \systemNM{} & 35.2 & 24.0 & 7.6 & 33.2 & 59.2 & 59.5 & 82.2 & 69.0 \\
\quad \systemMM{} & \textbf{28.8} & 29.6 & \textbf{6.4} & 35.2 & 58.4 & 49.3 & \textbf{81.8} & \textbf{61.5} & \quad \systemMM{} & \textbf{38.4} & 27.6 & \textbf{5.4} & 28.6 & \textbf{66.0} & 58.2 & \textbf{87.7} & \textbf{69.9} \\
\quad \systemSM{} & 28.0 & 34.0 & 7.6 & 30.4 & \textbf{62.0} & 45.2 & 78.7 & 57.4 & \quad \systemSM{} & 36.0 & 27.8 & 6.8 & 29.4 & 63.8 & 56.4 & 84.1 & 67.5 \\
\midrule
\textbf{Qwen2.5-3B} &  &  &  &  &  &  &  &  & \textbf{Qwen2.5-7B} &  &  &  &  &  &  &  &  \\
\rowcolor[gray]{0.85}\quad \textit{Base} & \textit{28.0} & \textit{54.6} & \multicolumn{2}{c}{(\textit{17.4})} & \textit{82.6} & \textit{33.9} & - & - & \quad \textit{Base} & \textit{32.2} & \textit{61.6} & \multicolumn{2}{c}{(\textit{6.2})} & \textit{93.8} & \textit{34.3} & - & - \\
\quad \systemKP{} & 9.0 & \textbf{8.0} & 18.0 & \textbf{65.0} & 17.0 & \textbf{52.9} & 33.3 & 40.9 & \quad \systemKP{} & 17.4 & \textbf{13.4} & 14.4 & \textbf{54.8} & 30.8 & \textbf{56.5} & 54.7 & 55.6 \\
\quad \systemNM{} & 17.8 & 19.8 & 11.4 & 51.0 & 37.6 & 47.3 & 61.0 & 53.3 & \quad \systemNM{} & 23.0 & 23.2 & 9.4 & 44.4 & 46.2 & 49.8 & 71.0 & 58.5 \\
\quad \systemMM{} & 18.8 & 21.6 & 9.4 & 50.2 & 40.4 & 46.5 & 66.7 & \textbf{54.8} & \quad \systemMM{} & \textbf{24.4} & 23.6 & \textbf{8.4} & 43.6 & 48.0 & 50.8 & \textbf{74.4} & \textbf{60.4} \\
\quad \systemSM{} & \textbf{19.8} & 29.6 & \textbf{7.6} & 43.0 & \textbf{49.4} & 40.1 & \textbf{72.3} & 51.6 & \quad \systemSM{} & \textbf{24.4} & 24.4 & 9.2 & 42.0 & \textbf{48.8} & 50.0 & 72.6 & 59.2 \\
\midrule
\textbf{Llama-3.2-3B} &  &  &  &  &  &  &  &  & \textbf{Llama-3.1-8B} &  &  &  &  &  &  &  &  \\
\rowcolor[gray]{0.85}\quad \textit{Base} & \textit{35.8} & \textit{47.2} & \multicolumn{2}{c}{(\textit{17.0})} & \textit{83.0} & \textit{43.1} & - & - & \quad \textit{Base} & \textit{37.4} & \textit{44.8} & \multicolumn{2}{c}{(\textit{17.8})} & \textit{82.2} & \textit{45.5} & - & - \\
\quad \systemKP{} & 21.6 & \textbf{21.4} & 12.4 & \textbf{44.6} & 43.0 & \textbf{50.2} & 63.5 & 56.1 & \quad \systemKP{} & 25.8 & \textbf{20.8} & 12.4 & \textbf{41.0} & 46.6 & \textbf{55.4} & 67.5 & 60.9 \\
\quad \systemNM{} & \textbf{31.4} & 35.8 & \textbf{6.2} & 26.6 & \textbf{67.2} & 46.7 & \textbf{83.5} & \textbf{59.9} & \quad \systemNM{} & \textbf{35.6} & 38.6 & 4.4 & 21.4 & \textbf{74.2} & 48.0 & 89.0 & 62.4 \\
\quad \systemMM{} & 28.4 & 33.6 & 8.2 & 29.8 & 62.0 & 45.8 & 77.6 & 57.6 & \quad \systemMM{} & 35.0 & 36.2 & 5.2 & 23.6 & 71.2 & 49.2 & 87.1 & \textbf{62.8} \\
\quad \systemSM{} & 28.6 & 33.4 & 6.8 & 31.2 & 62.0 & 46.1 & 80.8 & 58.7 & \quad \systemSM{} & 34.2 & 38.6 & \textbf{3.8} & 23.4 & 72.8 & 47.0 & \textbf{90.0} & 61.7 \\
\midrule
\textbf{Gemma-3-4b} &  &  &  &  &  &  &  &  & \textbf{Gemma-3-12b} &  &  &  &  &  &  &  &  \\
\rowcolor[gray]{0.85}\quad \textit{Base} & \textit{33.4} & \textit{66.6} & \multicolumn{2}{c}{(\textit{-})} & - & \textit{33.4} & - & - & \quad \textit{Base} & \textit{44.0} & \textit{56.0} & \multicolumn{2}{c}{(\textit{-})} & - & \textit{44.0} & - & - \\
\quad \systemKP{} & 19.4 & \textbf{14.8} & 14.0 & \textbf{51.8} & 34.2 & \textbf{56.7} & 58.1 & 57.4 & \quad \systemKP{} & 26.4 & \textbf{15.0} & 17.2 & \textbf{41.4} & 41.4 & \textbf{63.8} & 60.5 & 62.1 \\
\quad \systemNM{} & 23.2 & 18.8 & 10.0 & 48.0 & 42.0 & 55.2 & 69.9 & 61.7 & \quad \systemNM{} & \textbf{36.4} & 27.0 & \textbf{7.0} & 29.6 & 63.4 & 57.4 & \textbf{83.9} & 68.2 \\
\quad \systemMM{} & \textbf{25.6} & 27.0 & \textbf{7.4} & 40.0 & \textbf{52.6} & 48.7 & \textbf{77.6} & 59.8 & \quad \systemMM{} & 35.6 & 29.6 & \textbf{7.0} & 27.8 & \textbf{65.2} & 54.6 & 83.6 & 66.0 \\
\quad \systemSM{} & 25.4 & 19.8 & 9.0 & 45.8 & 45.2 & 56.2 & 73.8 & \textbf{63.8} & \quad \systemSM{} & 35.2 & 24.6 & 7.6 & 32.6 & 59.8 & 58.9 & 82.2 & \textbf{68.6} \\
\bottomrule
\end{tabular}
\end{table*}

\section{MMLU Benchmark results}
\label{sec:app:benchark}
We show in Table~\ref{tab:mmlu} the results of a world knowledge benchmark, MMLU~\cite{hendryckstest2021}, on the trained models including \systemKP{}, \systemNM{}, \systemMM{} and \systemSM{}. The results show almost intact behavior, indicating that the hallucination verification training did not lead to model collapse.  

\begin{table}[!htp]\centering
\small
\caption{Benchmarking trained models on world knowledge}\label{tab:mmlu}
\label{tab:benchmarks}
\begin{tabular}{lr |lr}\toprule
Model &MMLU & Model &MMLU\\\midrule

\textbf{Gemma-3-12b-it} & &\textbf{Llama-3.1-8B-Instruct} & \\
\cellcolor[HTML]{DDDDDD}\textit{Base} &\cellcolor[HTML]{DDDDDD}71.51\% &\cellcolor[HTML]{DDDDDD}\textit{Base} &\cellcolor[HTML]{DDDDDD}68.22\% \\
\systemKP{} &71.34\% & \systemKP{} &67.39\% \\

\systemSM{} &70.81\% & \systemSM{} &66.71\% \\

\systemMM{} &70.90\% & \systemMM{} &66.73\% \\

\systemNM{} &70.81\% & \systemNM{} &66.64\% \\
\midrule
\textbf{Gemma-3-4b-it} & & \textbf{Llama-3.2-3B-Instruct} & \\
\cellcolor[HTML]{DDDDDD}\textit{Base} &\cellcolor[HTML]{DDDDDD}57.63\% &\cellcolor[HTML]{DDDDDD}\textit{Base} &\cellcolor[HTML]{DDDDDD}62.26\% \\
\systemKP{} &57.47\% &\systemKP{} &61.47\% \\
\systemSM{} &57.25\% &\systemSM{} &61.13\% \\
\systemMM{} &57.34\% &\systemMM{} &61.10\% \\
\systemNM{} &57.14\% &\systemNM{} &60.80\% \\
\midrule

\textbf{Gemma-2-2b-it} & &\textbf{Qwen2.5-3B-Instruct} &\\

\cellcolor[HTML]{DDDDDD}\textit{Base} &\cellcolor[HTML]{DDDDDD}56.84\% &\cellcolor[HTML]{DDDDDD}\textit{Base} &\cellcolor[HTML]{DDDDDD}65.50\% \\

\systemKP{} &56.88\% &\systemKP{} &65.32\% \\

\systemSM{} &56.40\% &\systemSM{} &65.41\% \\

\systemMM{} &56.45\% &\systemMM{} &65.43\% \\

\systemNM{} &56.52\% &\systemNM{} &65.39\% \\
\midrule

\textbf{Gemma-2-9b-it} & &\textbf{Qwen2.5-7B-Instruct} & \\

\cellcolor[HTML]{DDDDDD}\textit{Base} &\cellcolor[HTML]{DDDDDD}71.87\% &\cellcolor[HTML]{DDDDDD}\textit{Base} &\cellcolor[HTML]{DDDDDD}71.73\% \\

\systemKP{} &71.57\% &\systemKP{} &71.74\% \\

\systemSM{} &71.21\% &\systemSM{} &72.01\% \\

\systemMM{} &71.61\% &\systemMM{} &72.10\% \\

\systemNM{} &71.25\% &\systemNM{} &71.95\% \\

\bottomrule
\end{tabular}
\end{table}

\end{document}